\documentclass{article}

\PassOptionsToPackage{numbers,compress}{natbib}
\usepackage[preprint]{neurips_2024}

\usepackage{wrapfig}
\usepackage[utf8]{inputenc} % allow utf-8 input
\usepackage[T1]{fontenc}    % use 8-bit T1 fonts
\usepackage{hyperref}       % hyperlinks
\usepackage{url}            % simple URL typesetting
\usepackage{booktabs}       % professional-quality tables
\usepackage{pifont}         % for \ding symbols
\usepackage{amsfonts}       % blackboard math symbols
\usepackage{nicefrac}       % compact symbols for 1/2, etc.
\usepackage{microtype}      % microtypography
\usepackage[table]{xcolor}
\usepackage{lipsum}
\usepackage{graphicx}
\usepackage{times}
\usepackage{amsmath,amssymb}
\usepackage{enumitem}
\usepackage{xspace}
\usepackage{pifont} % for the \ding
\usepackage{multirow, makecell, tabularx}
\newcolumntype{C}{>{\centering\arraybackslash}X}
\newcolumntype{W}{>{\hsize=1.1\hsize\centering\arraybackslash}X} % wider

\usepackage{subcaption}
\usepackage{algorithm}
\usepackage{algpseudocode}

\newcommand{\method}{GEM-2\xspace}
\newcommand{\methodtwo}{GEM-2.1\xspace}

\newcommand{\verify}[1]{\textcolor{red}{Verify.}}

\newcommand{\cmark}{\ding{51}}%
\definecolor{lightgray}{gray}{0.75}
\newcommand{\xmark}{\textcolor{lightgray}{\ding{55}}}% X mark symbol
\definecolor{pastelblue}{RGB}{70, 70, 70} % Light Blue

\usepackage{booktabs,tabularx}
\usepackage[most]{tcolorbox}
\definecolor{greydark}{RGB}{90, 90, 90} 
\definecolor{greylight}{RGB}{245, 245, 245}

\newtcolorbox{greycustomblock}{
  colframe=greydark,        % Left line color
  colback=greylight,        % Background color
  boxrule=1pt,              % Line width
  left=2.5pt,               % Inner left margin
  right=3pt,                % Inner right margin
  top=5pt,                  % Inner top margin
  bottom=3pt,               % Inner bottom margin
  arc=0pt,                  % No rounded corners
  breakable,                % Allow breaking across pages
  before skip=0.2\baselineskip, % Vertical space before
  after skip=0.2\baselineskip,  % Vertical space after
  left skip=0pt,            % Left skip (adjust as needed)
  right skip=0pt,           % Right skip (adjust as needed)
  enhanced jigsaw,          % Enhanced jigsaw for precise control
  frame hidden,             % Hide the default frame
  overlay={                 % Custom overlay for left line
    \draw[greydark, line width=2pt]
      ([yshift=-1pt]frame.north west) -- ([yshift=1pt]frame.south west); % Adjusting the y coordinates to match exactly
  },
  fontupper=\selectfont, % Font size and style for the table
}

\newtcolorbox{roundgrey}{
  colframe=greydark,        % Frame color
  colback=greylight,        % Background color
  boxrule=0.8pt,            % Frame line width
  left=5pt,                 % Inner left margin
  right=5pt,                % Inner right margin
  top=5pt,                  % Inner top margin
  bottom=5pt,               % Inner bottom margin
  arc=4pt,                  % Rounded corners
  breakable,                % Allow breaking across pages
  before skip=0.2\baselineskip,
  after skip=0.2\baselineskip,
  left skip=0pt,
  right skip=0pt,
  enhanced,                 % Enable advanced styling
  frame hidden=false,       % Show normal frame
  fontupper=\selectfont,    % Normal font inside
}

\usepackage[most]{tcolorbox}
\usepackage{titling} % lets us suppress the normal \maketitle if we want
\usepackage{setspace} % for slight looseness if you want

\definecolor{frontbg}{RGB}{247,249,252}      % big box fill
\definecolor{frontborder}{RGB}{210,215,225}  % big box border
\definecolor{fronttext}{RGB}{20,20,20}       % near-black text
\definecolor{frontsub}{RGB}{90,90,90}        % subtitle/affiliation grey

\newtcolorbox{frontpagebox}{
  breakable,
  colback=frontbg,
  colframe=frontborder,
  boxrule=0.5pt,
  arc=10pt,              % <-- rounded corners
  left=16pt,
  right=16pt,
  top=16pt,
  bottom=16pt,
  before skip=0pt,
  after skip=1.5em,
  enhanced,
}

\newtcolorbox{metadatabox}{
  colback=white,
  colframe=frontborder,
  boxrule=0.5pt,
  arc=8pt,
  left=10pt,
  right=10pt,
  top=8pt,
  bottom=8pt,
  enhanced,
}

\usepackage{fontawesome}
\usepackage{adjustbox} % Add this line to your preamble
\usepackage{tikz}

\newcommand{\lightbulbicon}{%
  \begin{tikzpicture}[baseline=-0.5ex]
    \draw[fill=white, draw=insightteal, thick] (0,0) circle (1.5ex);
    \node[scale=0.8, color=insightteal] at (0,0) {\faLightbulbO~};
  \end{tikzpicture}%
}

\definecolor{insightteal}{RGB}{34, 139, 139}   % A sophisticated, medium-dark teal
\definecolor{insightback}{RGB}{240, 248, 248}   % A very light, complementary cyan-white

\newtcolorbox{customblockquote}{
  colframe=insightteal,
  colback=insightback,
  boxrule=0pt,
  left=5pt,  % Set to 0pt so the background color touches the left 
  right=4pt,
  top=5pt,
  bottom=3pt,
  arc=0pt,
  breakable,
  before skip=1.2\baselineskip,
  after skip=0.7\baselineskip,
  left skip=0pt,
  right skip=0pt,
  enhanced jigsaw,
  frame hidden,
   overlay={
    \draw[insightteal, line width=2pt] 
      ([yshift=1pt]frame.north west) -- (frame.south west);
    \node[inner sep=0pt] at ([xshift=0pt, yshift=-1.3pt]frame.north west) {\lightbulbicon};
  },
  fontupper=\fontfamily{lmr}\selectfont,
  boxsep=1pt,
}

\newcommand{\fronttitle}[1]{%
  {\sffamily\bfseries\LARGE #1\par}
}

\newcommand{\frontauthors}[1]{%
  {\sffamily\normalsize\textbf{#1}\par}
}

\newcommand{\frontabstracttext}[1]{%
  {\small #1\par}
}

\usepackage{titletoc}

\titlecontents{section}[0em]
{\addvspace{0.3em}\sffamily\bfseries\small\color{fronttext}}
{\thecontentslabel\hspace{1em}}
{\hspace*{-1em}}
{\titlerule*[0.7pc]{\textcolor{frontborder}{.}}\thecontentspage}

\titlecontents{subsection}[2.5em]
{\addvspace{0pt}\sffamily\footnotesize\color{greydark}}
{\thecontentslabel\hspace{1em}}
{\hspace*{-1em}}
{\titlerule*[0.7pc]{\textcolor{frontborder}{.}}\thecontentspage}

\newtcolorbox{executiveTOC}{
  colback=white,
  colframe=frontborder,
  boxrule=0.5pt,
  arc=6pt,
  left=20pt, right=20pt, top=14pt, bottom=14pt,
  before skip=1em,
  after skip=2em,
  enhanced,
}
\newenvironment{frontabstract}{\setlength{\parindent}{0pt}}{}

\begin{document}

\thispagestyle{empty}  % no header/footer number on first page

\begin{frontpagebox}
  \begin{frontabstract}

    % --- Title
    \fronttitle{Probabilistic Transformers for Joint \\ Modeling of Global Weather Dynamics and Decision-Centric Variables}

    \vspace{0.6em}

    % --- Authors / Team line
    \frontauthors{Salient}

    {\sffamily\footnotesize\color{frontsub}%
      Paulius Rauba$^{\star}$\footnote{University of Cambridge \quad $^{\star}$Equal contribution}, Viktor Cikojevi\'{c}$^{\star}$, Fran Bartoli\'{c}$^{\star}$, Sam Levang$^{\star}$, Ty Dickinson, 
      Chase Dwelle\\
    \par}

    \vspace{1.2em}

    % --- Execute-summary paragraph 
    \frontabstracttext{
    \vspace{-5mm}

 Weather forecasts sit upstream of high-stakes decisions in domains such as grid operations, aviation, agriculture, and emergency response. Yet forecast users often face a difficult trade-off. Many decision-relevant targets are functionals of the atmospheric state variables, such as extrema, accumulations, and threshold exceedances, rather than state variables themselves. As a result, users must estimate these targets via post-processing, which can be suboptimal and can introduce structural bias. The core issue is that decisions depend on distributions over these functionals that the model is not trained to learn directly.
    
    $\space$

    In this work, we introduce \method, a probabilistic transformer that jointly learns global atmospheric dynamics alongside a suite of variables that users directly act upon. 
    Using this training recipe, we show that a lightweight ($\sim$275M params) and computationally efficient ($\sim$20-100x training speedup relative to state-of-the-art) transformer trained on the CRPS objective can directly outperform operational numerical weather prediction (NWP) models and be competitive with ML models that rely on expensive multi-step diffusion processes or require bespoke multi-stage fine-tuning strategies. We further demonstrate state-of-the-art economic value metrics under decision-theoretic evaluation, stable convergence to climatology at S2S and seasonal timescales, and a surprising insensitivity to many commonly assumed architectural and training design choices.

    \vspace{0.1em}

    }

    % --- metadata card, like Meta's "Date / Code / Weights ..."
    \begin{metadatabox}
      {\footnotesize
        \textbf{Date:} January, 2026\\
        \textbf{Contact:} \texttt{\{name.surname\}@salient.com}
      }
    \end{metadatabox}

  \end{frontabstract}
\end{frontpagebox}

\begin{executiveTOC}
  {\sffamily\bfseries\footnotesize\textcolor{insightteal}{CONTENTS}}
  \vspace{0.6em}
  
  % Reset depth to show Sections (1) and Subsections (2) for the main body
  \setcounter{tocdepth}{2} 

  \makeatletter
  \renewcommand\tableofcontents{\@starttoc{toc}}
  \makeatother
  
  {
    \hypersetup{linkcolor=black}
    \tableofcontents
  }
\end{executiveTOC}

\newpage

\section{Introduction}
\label{sec:intro}

How can we build forecasting systems that are optimal for end-users of weather forecasts? This is the primary question we consider in this work.

%\subsection*{Machine learning models that optimize forecast skill}

Accurate and reliable weather forecasts underpin safety, planning, and economic activity. Recent machine learning (ML) models now match or surpass leading numerical weather prediction (NWP) systems on research benchmarks on short and medium-range timescales \citep{pathak2022fourcastnet, bi2023pangu, chen2023fuxi, lam2023learning, chen2025fengwu, bonev2025fourcastnet3}. While most early ML systems were deterministic, the weather community relies on probabilistic \emph{ensembles} (e.g., ECMWF ENS, NOAA GEFS) to quantify forecast uncertainty and support risk-aware decision-making \citep{documentation2020part, hamill2022reanalysis}. In parallel, ML-based probabilistic forecasters have advanced quickly, using diffusion \citep{price2023gencast, price2024probabilistic}, flow-matching \citep{couairon2024archesweather}, CRPS-trained ensembles \citep{alet2025skillful, lang2024aifs}, and hybrid ML--physics designs \citep{kochkov2024neuralgcm, sun2025fuxiweather}, closing the gap to NWP ensembles and, in several cases, outperforming them on headline metrics.

Most of these models optimize \emph{forecast skill}. However, end users rarely act on the raw predictions of a model. Instead, they are interested in decision-centric targets (e.g., daily minima or maxima). This focus area has been relatively neglected but is increasing in importance \citep{raeth2025evaluating}. We refer to these targets as \textit{diagnostic variables}, since they are the quantities end users use to make decisions. For instance, daily temperature extremes drive energy demand and correlate strongly with electricity load, so utilities rely on estimates of daily extremes for planning \citep{li2021spatial, wang2025short}. 
Wind gusts are another example, since many impact models respond nonlinearly to wind speed and are driven by peak gusts rather than mean wind.

Therefore, we aim to build \textit{actionable} weather forecasting models, meaning models whose end-to-end pipeline supports decision-making directly. This setting imposes additional, orthogonal requirements: (i) \emph{decision-making variables} (e.g., daily aggregates) that users act on directly; (ii) \emph{throughput and latency} constraints that enable large ensembles, fast event fine-tuning, and cheap reforecasts; and (iii) long-lead behavior consistent with climatology, including convergence toward climatological baselines.
\vspace{-2mm}
\subsection*{Global weather forecasting with native decision-centric outputs}

In this work, we introduce a decision-aligned probabilistic forecaster \method that addresses these three criteria. Our contributions are three-fold.

\textbf{First}, we formalize the decision-aligned forecasting problem. We separate the targets into prognostic variables, which drive the autoregressive evolution of the atmospheric state, and diagnostic variables, which are forecasted explicitly but not fed back into the state. This formulation yields a clear utility-theoretic interpretation: if post-processing the prognostic variables results in worse decision-centric variables, then directly modeling those variables should improve end user decisions. 
The primary mechanism is that the variables that drive the dynamics of the atmosphere are often different from the variables of interest, and \textit{jointly} optimizing them can yield superior decisions for users of weather forecasts.

\textbf{Second}, motivated by this insight, we construct \method, a computationally efficient global transformer that jointly models prognostic and diagnostic variables. \method introduces \emph{periodic shifted-window attention}, which adapts Swin-style locality to the spherical topology by enforcing periodicity along longitude and non-periodicity along latitude. The model uses a single-shot probabilistic architecture trained using a strictly proper composite objective. It is decision-centric: daily aggregates and other user-facing diagnostics are emitted natively, jointly optimized with the prognostic trajectories. This multi-target supervision reduces the impedance mismatch between model outputs and downstream decision pipelines, unlike approaches that require separate post-processing stages.

\textbf{Third}, we provide an extensive empirical investigation spanning operational ensembles, and multiple ablation axes. While \method is lightweight ($\sim$275M params) and cheap ($\sim$20-100x training speedup relative to SOTA), we outperform computationally heavy operational models and are competitive with much larger models that have expensive diffusion-oriented denoising stages or complicated multi-stage fine-tuning strategies. We show that \method produces state-of-the-art performance on probabilistic metrics measured against reanalysis and station observations across 1–126 day leads, with stable convergence toward climatology at subseasonal to seasonal timescales. We further conduct a broad suite of ablations at 1° resolution, varying conditioning schemes, Markov order, grid representation, depth, embedding dimension, number of prognostic vertical levels, rollout strategy, and the presence or absence of spectral regularization. Contrary to common assumptions, we find that our model is extremely stable across many design choices. For instance, the performance does not change with moving to different representational spaces, employing multi-lead CRPS fine-tuning or increasing the Markov order.

Taken together, our results provide that a decision-aligned, computationally parsimonious transformer architecture that can achieve high-quality probabilistic forecasts without relying on more custom spherical architectures. Table~\ref{tab:forecasting_models} situates \method within the existing ecosystem of ML weather models. The remainder of the paper develops the theoretical framing, architectural components, scoring objectives, and empirical findings in detail.

\begin{figure}
    \centering
    \includegraphics[trim={2cm 3.5cm 4.5cm 2cm},clip, width=\linewidth]{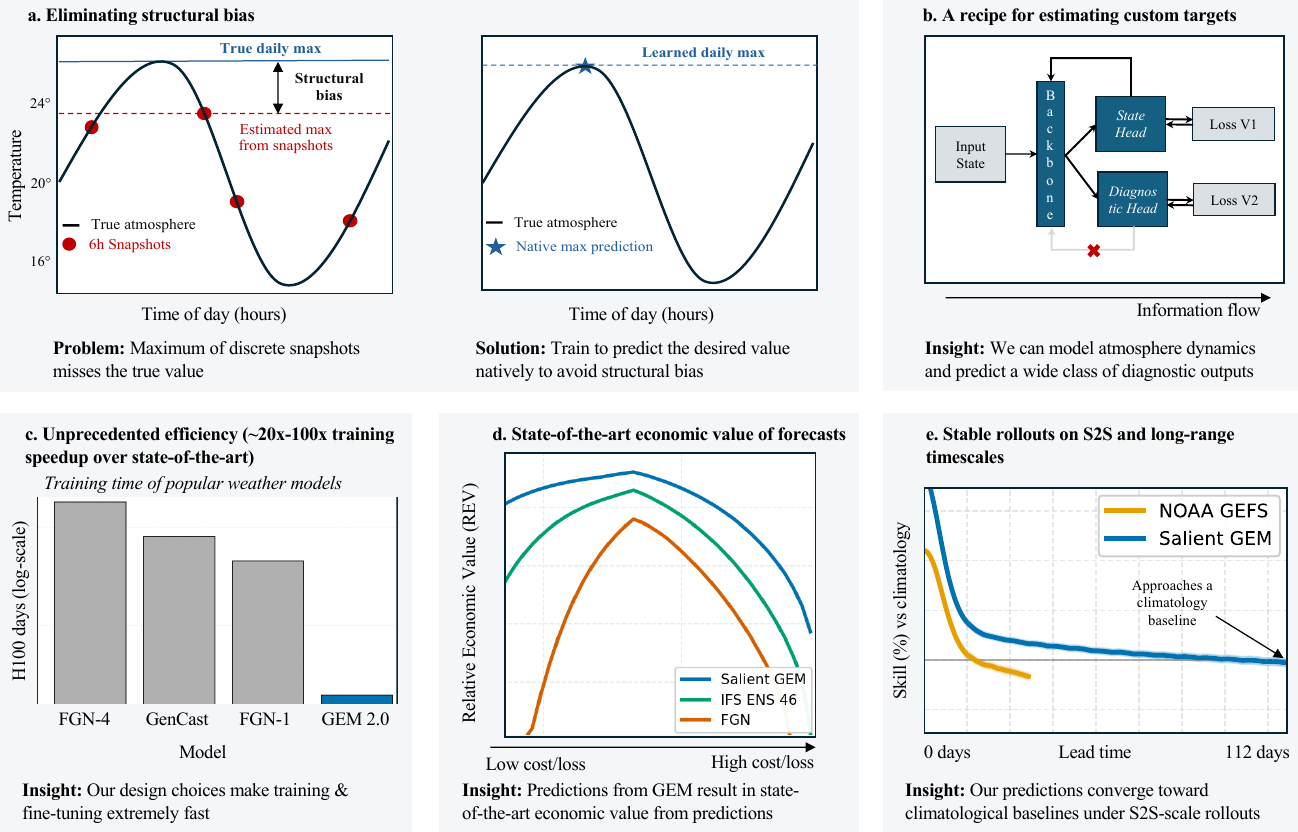}
    \caption{\textbf{Highlights of this work}. \textbf{(a)} We show that modeling decision-centric variables natively eliminates structural bias which arises from post-hoc aggregation of snapshots; \textbf{(b)} We introduce a fine-tuning recipe that allows us to model variables of interest natively for different classes of diagnostic variables; \textbf{(c)}  We obtain extreme efficiency, with $\sim$100x reduction in training cost and $\sim$10x reduction in inference costs relative to SOTA models; \textbf{(d)} Our predictions demonstrate state-of-the-art economic value for end-users of predictions of forecasts; \textbf{(e)} We obtain stability for free within our modeling paradigm with long-lead convergence to climatology. This stability is manifested as convergence toward climatological baseline in the long-lead domain \textit{Note}: FGN-1 and FGN-4 refer to the 1-seed and 4-seed model of FGN, respectively.}
    \label{fig:fig1main}
\end{figure}

\begin{table*}[t]
\centering
\scriptsize
\renewcommand{\arraystretch}{1.2}
\begin{tabularx}{\textwidth}{@{}l c c c c c c c c@{}}
\toprule
\textbf{Model} & \textbf{Input times} & \textbf{Step} & \textbf{Res.} & \textbf{\# Vars} & \textbf{Max Lead} & \textbf{Type} & \textbf{\# Params} & \textbf{Diagnostics} \\
\midrule

\multicolumn{9}{c}{\textit{Deterministic ML Weather Models}} \\
\midrule
FourCastNet (v1/2) \citep{pathak2022fourcastnet} & $0$ & 6h & 0.25° & 20 & 10 d & Det. & $\sim$433M & \xmark \\
GraphCast \citep{lam2023learning} & $-6,0$ & 6h & 0.25° & 227 & 10 d & Det. & $\sim$37M & \xmark \\
Pangu-Weather \citep{bi2023pangu} & $0$ & 1/3/6/24h & 0.25° & 69 & 7 d & Det. & $\sim$256M & \xmark \\
FuXi \citep{chen2023fuxi} & $-6,0$ & 6h & 0.25° & 70 & 15 d & Det. & >3B & \xmark \\
FengWu \citep{chen2025fengwu} & $-6,0$ & 6h & 0.25° & 69 & 10 d & Det. & n/a & \xmark \\
Keisler GNN \citep{keisler2022gnn} & $0$ & 6h & $\sim$1° & 78 & 6 d & Det. & 6.7M & \xmark \\
AIFS (Single) \citep{lang2024aifs} & $-6,0$ & 6h & 0.25° & $\sim$174 & 10 d & Det. & $\sim$229M & \xmark \\
ArchesWeather \citep{couairon2024archesweather} & $0$ & 24h & 1.5° & 82 & 15 d & Det. & 44M-84M & \xmark \\
\midrule

\multicolumn{9}{c}{\textit{Probabilistic ML Models}} \\
\midrule
GenCast \citep{price2023gencast} & $-12, 0$ & 12h & 0.25° & 84 & 15 d & Prob. & n/a & \xmark \\
FGN \citep{alet2025skillful} & $-6,0$ & 6h & 0.25° & 84 & 15 d & Prob. & $\sim$720M & \xmark \\
FourCastNet-3 \citep{bonev2025fourcastnet3} & $0$ & 6h & 0.25° & 84 & 60 d & Prob. & $\sim$710M & \xmark \\
AIFS-ENS/CRPS \citep{lang2024aifs} & $-6,0$ & 6h & $\sim$0.25° & n/a & 10 d & Prob. & $\sim$229M & \xmark \\
ArchesGen \citep{couairon2024archesweather} & -24, 0 & 24h & 1.5° & 82 & 15 d & Prob. & n/a & \xmark \\
DLWP \citep{weyn2021subseasonal} & $-6,0$ & 6h & 1.4° & 6 & 42 d & Prob. & $\sim$2.6M & \xmark \\
\rowcolor{gray!10}
\textbf{\method (this work)} & 0 & 24h & 0.25° & 56 & 126d & Prob. & 275M & \cmark \\
\midrule

\multicolumn{9}{c}{\textit{Hybrid, Assimilation, and Foundation Approaches}} \\
\midrule
FuXi Weather \citep{sun2025fuxiweather} & Cycled & 6h & 0.25° & --- & 10 d & Hybrid & n/a & \xmark \\
NeuralGCM \citep{kochkov2024neuralgcm} & 0 & 0.5h & Variable & 7 & Climate & Hybrid & $\sim$11-30M & \xmark \\
ClimaX \citep{nguyen2023climax} & 0 & 6h & Variable & 48 & $\sim$ 30d & Foundation & n/a & \xmark \\

\bottomrule
\end{tabularx}
\caption{\textbf{Technical specification of primary ML weather forecasting models.} Models are grouped by paradigm: deterministic, probabilistic, and hybrid/foundation approaches. Each entry summarizes temporal setup, spatial resolution, variables, lead time, model type, and size. Our best estimate of number of parameters is reported. For models which do not report explicitly, we either estimate it if it's possible to do so or omit entirely. Diagnostics refers to models which explicitly jointly model user-centric variables and global weather dynamics, excluding accumulated precipitation. {\footnotesize \textbf{\textcolor{pastelblue}{NOTE}}}: Max lead refers to the largest publicly reported max lead time. \# Params refers to the params of the final and/or largest reported model. }
\label{tab:forecasting_models}
\end{table*}

\vspace{-2mm}
\begin{customblockquote} 
\paragraph{Takeaway.} This report formulates ML weather forecasting as a decision-centric problem, introduces probabilistic transformers that natively emit user-facing diagnostic variables alongside prognostic atmospheric fields using boundary-aware attention mechanisms and spectral log-power CRPS, and demonstrates that this design  yields lightweight global models with state-of-the-art skill on decision-relevant variables and stable convergence at long lead times.
\end{customblockquote}

\section{Actionable forecasting as a decision problem}
\label{sec:decision}

In this work, we are interested in forecasts that are not only \emph{skillful}, but also \emph{actionable}: forecasts that can be used directly in downstream decision rules \citep{murphy1993good, katz1997economic}. Downstream users do not act on the full prognostic state of the atmosphere; instead, they act on a comparatively small set of derived quantities such as daily extrema, threshold exceedances, and related aggregates. Such derived quantities can even be defined on different spatial or temporal grids.

Most existing ML weather models are trained to predict prognostic fields consisting primarily of instantaneous atmospheric snapshots at the model timestep, and any sub-timestep or temporal aggregations must be inferred via post-processing. In contrast, our starting point is different:

\begin{quote}
    \emph{If decision-makers use diagnostic variables, then the forecasting system should model and emit those diagnostics directly.}
\end{quote}

The use of strictly proper scoring rules (such as CRPS) to train probabilistic forecasts is by now adopted in some weather forecasting works \citep[e.g.][]{alet2025skillful, lang2024crps}. What is new in our formulation is not the choice of score, but the object it is applied to: rather than primarily scoring the raw prognostic fields $X_t$ and deriving diagnostics afterwards, we explicitly construct and train a model whose primary output is the distribution of decision-relevant diagnostics $Y_t$ at each lead time. In this section, we formalize this viewpoint.

\subsection{Forecasting setup}

We work in a global probabilistic forecasting setting and employ the equirectangular (lat/lon) projection, with height $H_\text{grid}$ and width $W_\text{grid}$. At each lead time $t \ge 1$, we distinguish three types of fields:
\begin{itemize}
    \item \textbf{Prognostic fields} $X_t \in \mathbb{R}^{C_x \times H_\text{grid} \times W_\text{grid}}$: gridded atmospheric state variables, represented as hourly snapshots (e.g., temperature, winds, pressure), that define the system dynamics and are explicitly forecast forward in time.
    \item \textbf{Diagnostic targets} $Y_t \in \mathbb{R}^{C_y \times H_\text{grid} \times W_\text{grid}}$: user-facing quantities (such as daily minima and maxima), which enter decision rules but are not fed back into the state transition\footnote{the diagnostics in this work have the same spatial distribution as prognostic fields; but that's an assumption we can easily relax (see Sec. \ref{sec:typology})}.
    \item \textbf{Conditioning fields} $C \in \mathbb{R}^{C_c \times H_\text{grid} \times W_\text{grid}}$: static or slowly varying fields that constrain the evolution but are not themselves forecast.
\end{itemize}

Let $X_{\le 0}$ denote the assimilated history up to initialization time. We write
\(    H := (X_{\le 0}, C) \in \mathcal{H}
\)
for the information available to the forecaster at initialization, with $\mathcal{H}$ the corresponding state space. A particular forecast case corresponds to an observed realization $H = h$. Under the true atmospheric dynamics, for each lead time $t$ there exists an unknown conditional distribution of the diagnostics given this information,
\[
    p_t(\,\cdot \mid h) \;:=\; \mathbb{P}(Y_t \in \cdot \mid H = h),
    \qquad
    Y_t \mid H = h \sim p_t(\,\cdot \mid h).
\]
The forecasting system does not know $p_t$. Instead, it emits an approximate conditional law
\[
    q_{\theta,t}(X_t,Y_t \mid h) \;\approx\; p_t(X_t,Y_t \mid h),
\]
parameterized by network weights $\theta$. Thus, for each realized input $h$, the model provides a joint probabilistic forecast $q_{\theta,t}(X_t,Y_t \mid h)$ of the next-step state $X_t$ and the diagnostics $Y_t$. In our formulation, primary decision-relevant outputs of the system are distributions over $Y_t$; the evolution of the prognostic state $X_t$ is interesting in its own right and might be used for downstream applications, but we model it insofar as it is needed to support the diagnostic forecasts. 

\subsection{How do users interact with weather forecasts?}
End users rarely use the raw prognostic fields of a weather system. Instead, they rely on task-specific information which is required for decision-making. This information is typically derivative from the weather dynamics but is not the weather dynamics itself. Therefore, we want to provide end users with information that directly informs their use cases. Concretely, if a user takes an action $a_t \in \mathcal{A}_t$ on the basis of some variable $Y_t$, then we want to model $Y_t$ directly. This is because the actions of the user depend on the information provided, and can be represented by an overall utility $U_t(a_t, Y_t)$, where $U_t: \mathcal{A}_t \times \mathcal{Y}_t \to \mathbb{R}$. Here, $\mathcal{Y}_t$ is the domain of possible diagnostic realizations at time $t$. Therefore, with the forecast in place, the user can take the action that maximizes their expected utility for a decision with the information provided by the diagnostic forecast. Precisely, the user can solve:

\begin{equation}
    a_t^\ast
    \;=\;
    \arg\max_{a \in \mathcal{A}_t}
    \mathbb{E}_{Y_t \sim p_t(\cdot \mid H)}\big[\, U_t(a, Y_t) \,\big].
    \label{eq:bayes-opt-q}
\end{equation}

where \(p_t(\cdot \mid H)\) is the true conditional distribution of diagnostics given the history \(H\) at lead time \(t\). Relying on the prognostic states \(X_{\leq t}\) to approximate \(Y_t\) is often suboptimal because a sequence of forecasted state snapshots can be an inadequate statistic for a decision-relevant diagnostic. In practice, \(Y_t\) is typically a (possibly complex) functional of the underlying continuous-time atmospheric evolution and may depend on temporal extremes, subgrid variability, or additional domain-specific transformations; obtaining \(Y_t\) from \(X_{\leq t}\) can therefore require a separate post-processing or impact model, introducing additional assumptions and error. For instance, if end users care about the minimum daily temperature, the snapshots \(X_{\leq t}\) provide only a proxy for the true minimum temperature when they sparsely sample the diurnal cycle. In standard pipelines, \(Y_t\) is computed as a functional \(\phi\) of the forecasted state snapshots, yielding the estimator \(\hat{Y}_t = \phi(\{X_{t-1}, X_t\})\). However, because the snapshots represent a coarse sampling of the underlying diurnal dynamics, this estimator is generally biased: \(\mathbb{E}[\phi(X_{\leq t})\mid H] \neq \mathbb{E}[Y_t\mid H].\) More broadly, when \(\phi\) is nonlinear or requires an auxiliary model, the induced error can remain substantial even if the model captures the distribution of \(X_{\leq t}\) accurately. Thus, even if the model captures the distribution of \(X_{\leq t}\) perfectly, the resulting actions may yield lower expected utility than actions derived from a direct model of \(Y_t\).

Consider the case where $Y_t$ represents the daily minimum temperature. The true value is the minimum over a continuous interval, $Y_t = \min_{\tau \in [0,24]} \mathcal{X}(\tau)$,where $\mathcal{X}(\tau)$ denotes the temperature field at continuous time $\tau$ so that the discrete snapshot $X_t=\mathcal{X}(t)$. A standard forecaster producing discrete snapshots $\{X_k, \dots, X_t\}$ yields the estimator $\hat{Y}_t = \min \{X_k, \dots, X_t\}$. By definition, a minimum over a subset is greater than or equal to the minimum over the full domain, which introduces a systematic positive bias $\mathbb{E}[\hat{Y}_t] \geq \mathbb{E}[Y_t].$ This simple example demonstrates that forecasting a misaligned objective can impede modeling variables that end users care about. In fact, we demonstrate this behavior natively in Sec.~\ref{subsec:extreme_tails}, where we show that we can match or exceed modeling various diagnostic variables with fewer forward passes and smaller models relative to the state of the art. 

A good solution to the user-facing weather forecasting problem should therefore simultaneously satisfy two properties: \textbf{(a)} direct evolution of the variables which drive atmospheric dynamics; and \textbf{(b)} emission of desired user end-products.

\subsection{Why focus on emitting diagnostic variables directly?}

\begin{figure}
    \centering
    \includegraphics[trim={0.3cm 0.2cm 0.8cm 0},clip, width=\linewidth]{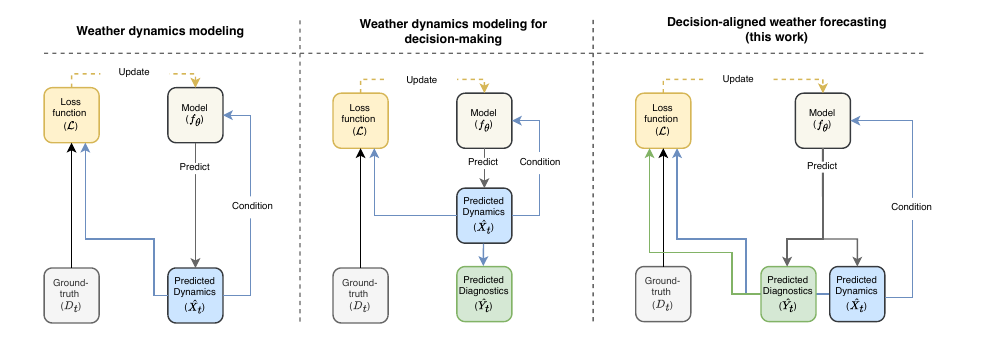}
    \caption{
    \textbf{From weather dynamics modeling to decision-aligned forecasting}. 
\textbf{Left}: Traditional forecasting frameworks propagate the full dynamical state forward in time by evolving prognostic variables that encode the system’s governing physics. 
\textbf{Middle}: Because these models provide only sparse state snapshots, downstream users often rely on post-processing to infer application-specific quantities. This workflow (i) ignores the structure of the desired targets during model training and (ii) introduces systematic bias when those targets cannot be reliably reconstructed from limited snapshots. 
\textbf{Right}: Decision-aligned forecasting integrates the evolution of weather states with the direct prediction of diagnostic targets—quantities of interest that do not influence the dynamics but are functionals of it. By coupling prognostic evolution with simultaneous diagnostic estimation, the approach yields the required outputs directly, eliminating dependence on post-hoc aggregation and reducing bias.}
    \label{fig:graph_dynamics}
\end{figure}

The formalism above makes precise a point that is intuitively clear to practitioners. In most operational settings, the role of a forecast is not to reconstruct the entire atmospheric state $X_t$ for its own sake, but to support decisions that depend on a comparatively low-dimensional collection of diagnostics $Y_t$ and simple functionals thereof. 

Conventional pipelines often reverse this priority: they train and evaluate primarily on instantaneous or sub-daily fields $X_t$, and only \emph{then} construct diagnostics by transforming model output. Here, we take the opposite view: we explicitly separate prognostic fields $X_t$, which drive the state evolution, from diagnostic targets $Y_t$, which enter decision rules, and we design the model to predict $Y_t$ natively at each lead time, rather than deriving it post-hoc. In practice, we neither observe utilities nor optimize expected utility directly. Instead, we train by minimizing the continuous ranked probability score. In a Bayesian context, scores are frequently referred to as utilities \citep{gneiting2007strictly}.

From an operational perspective, this separation between prognostic state and diagnostic targets also decouples the temporal resolution of the backbone from the effective temporal scale of the decisions. Because the model is trained to approximate the conditional laws \(p_t(Y_t \mid X_t, C)\) directly, quantities that depend on near-instantaneous behavior—daily maxima and minima, gust-like wind metrics, or short-window accumulations, we can avoid reconstructing these variables from sequences of sub-daily snapshots.  We can do this by learning a map from coarsely sampled trajectories \(X_t\) to the full distribution of daily extremes. This makes it possible to use comparatively coarse prognostic time steps (24\,h in our main system; cf.\ Section~\ref{sec:gem}) while retaining competitive skill on fast diagnostics, and it explains why GEM can match or exceed GenCast and FGN on tail-focused scores (Figures~\ref{fig:extreme_skill} and~\ref{fig:scores_ai}) with at least 4\(\times\) fewer forward passes than 6-hourly or diffusion-based generators (Table~\ref{tab:prob_models_compact_cited}).

Additionally, the use of a 24\,h timestep eliminates the need to accurately model the diurnal cycle in the prognostic state. We hypothesize that by removing this source of error accumulation the autoregressive rollouts, the coarse timestep is actually a key contributor to the rollout stability of GEM. Coupled together with the diagnostic framework, this configuration provides "stability for free" while still supporting the decision-aligned objective in Section~\ref{sec:decision}.

Two consequences follow from this discussion. First, any system that is trained primarily on $X_t$ and derives $Y_t$ via a fixed post-processing map $\phi$ is implicitly optimizing a surrogate problem that can be misaligned with Eq \eqref{eq:bayes-opt-q}, even if the surrogate achieves excellent forecast skill on snapshots. Second, once utilities are expressed in terms of diagnostics $Y_t$, it becomes natural to treat $p_t(Y_t \mid H)$ as the primary modeling target and to regard the evolution of $X_t$ as a mechanism that supports those diagnostic distributions. The factorization in Section~\ref{sec:gem} adopts exactly this perspective: the architecture is designed so that the learning objective is applied directly to $Y_t$, while still preserving a physically meaningful transition model for $X_t$.

\subsection{Typology of diagnostics}
\label{sec:typology}

\begin{figure}
    \centering
    \includegraphics[trim={1.2cm 0.2cm 0.1cm 0cm},clip, width=\linewidth]{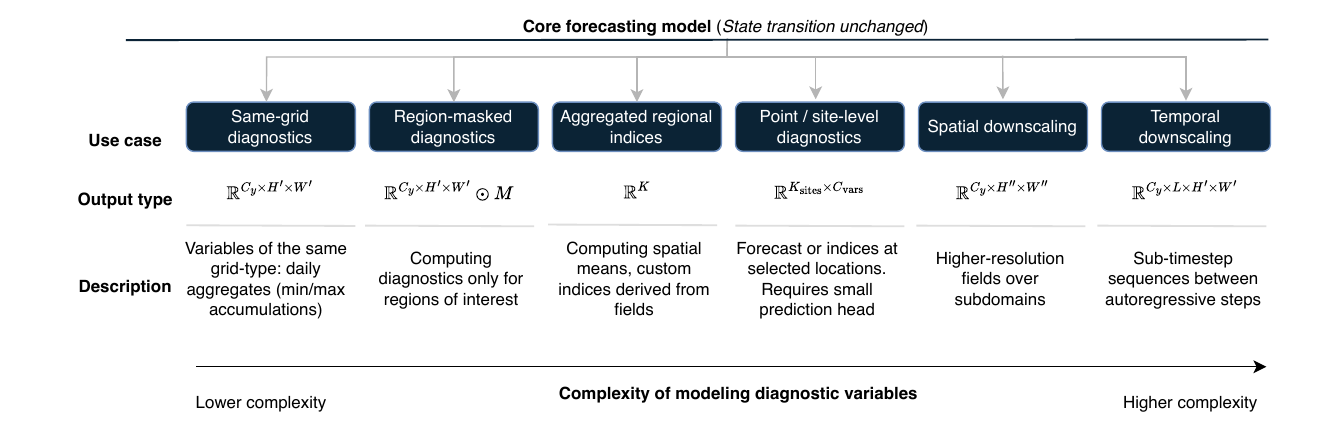}
\caption{\textbf{Typology of diagnostic targets.} Each class extends the forecasting system along a different structural axis—spatial, temporal, or aggregative—while leaving the underlying state transition unchanged. The primary mechanism of modeling weather states and optimizing the required user end-products at any spatial or temporal resolution jointly with the variables that drive the dynamics of the weather is the unifying component of all approaches.}
    \label{fig:diagnostics_subtypes}
\end{figure}

A decision-aligned forecaster must emit quantities that downstream users act upon. In practice, these diagnostic targets vary widely in structure: some live on the same grid as prognostic fields, others correspond to aggregates over regions or sites, and yet others represent refined temporal or spatial resolutions not present in the core autoregressive state. Although heterogeneous, these use cases share a common requirement: each diagnostic introduces an additional conditional distribution the model must represent, but none should alter the underlying transition dynamics. This leads to a simple organizing principle: diagnostics are side tasks defined on top of the prognostic trajectory, trained jointly through auxiliary proper scores, and parameterized so as not to interfere with the Markovian evolution of the state.

Figure ~\ref{fig:diagnostics_subtypes} summarizes the main diagnostic classes. The categories are arranged by the degree of structural departure from the native grid: from same-grid physical quantities (e.g.\ daily aggregates) to region-weighted functionals, point-level predictions, and various downscaling regimes. Importantly, most diagnostics require no additional transition parameters—only readout heads or masked losses. In this sense, the architecture learns a multitask mapping whose primary purpose is to expose the conditional laws that users act on, rather than to reshape the dynamics themselves. This maintains consistency with the overarching goal of matching the decision-relevant distribution of $Y_t$ while preserving parsimony in the forecasting operator.

Viewed through this lens, diagnostics serve as an energy-efficient extension of the forecasting operator: they widen the family of conditional distributions the model must approximate, yet leave the transition kernel untouched.

In the remainder of the paper we instantiate only a small subset of these diagnostic classes—primarily same-grid daily aggregates and simple regional functionals—but the typology fixes how more complex products should be integrated. Any new diagnostic corresponds to an additional conditional law of the form $p_t(Y_t \mid X_t, C)$ and therefore enters the system through emission heads and auxiliary scores. Section~\ref{sec:gem} makes this separation explicit in the modeling of $q_\theta(X_{1:T}, Y_{1:T} \mid X_{\le 0}, C)$, and Section~\ref{sec:exps} evaluates \method on both grid-based diagnostics and spatially aggregated quantities to verify that this design behaves as intended.

\section{\method: A probabilistic transformer for diagnostic modeling}
\label{sec:gem}

We instantiate the forecasting setup in Section~\ref{sec:decision} with  lightweight probabilistic transformers that are trained to emit user-facing products directly: \method. The model follows an FGN-style \citep{alet2025skillful} single-shot design, operates under the first-order Markov factorization, and jointly produces prognostic fields $X_t$ and diagnostic targets $Y_t$ at each lead.

Operationally, \method is our primary global system. It is trained on a $0.25^\circ$ grid and deployed for routine global forecasting, including a multi-year reforecast. In the course of deployment, we observed a small set of systematic prediction artifacts and remaining optimization degrees of freedom. \methodtwo is a successor architecture constructed to address these issues explicitly while keeping the learning objective fixed. Consequently, all $0.25^\circ$ headline results are reported for \method, whereas \methodtwo is used on a $1^\circ$ aggregated grid to isolate architectural changes and quantify their impact via targeted ablations.

\subsection{Forecasting factorization and objective}
In contrast to recent prior work, we operate under a different learning objective which aims to jointly learn the variables required for evolving atmospheric states and diagnostic outputs used by end users.

\begin{greycustomblock}
\textbf{Learning objective}.  Given the assimilated history $X_{\leq 0}$ and conditioning information $C$, our goal is to model $q_{\theta}(X_{1:T}, Y_{1:T}|X_{\leq 0}, C)$ as 
\begin{equation}
\label{eq:learning_objective}
\begin{aligned}
q_{\theta}(X_{1:T}, Y_{1:T} \mid X_{\leq 0}, C)
&= \prod_{t=1}^T q_{\theta}(Y_t, X_t \mid X_{\leq t-1}, C) \\
&= \prod_{t=1}^T
   \underbrace{q_{\theta}(X_t \mid X_{\leq t-1}, C)}_{\text{evolve state}}
   \underbrace{q_{\theta}(Y_t \mid X_{\leq t}, C)}_{\text{emit end products}}
\end{aligned}
\end{equation}

\end{greycustomblock}

We assume that diagnostics are side outputs that do not introduce temporal feedback, i.e., for all $t$, $(X_t, Y_t) \perp Y_{< t} | (X_{\leq t-1}, C)$. The factorization in \eqref{eq:learning_objective} separates two roles that the model must play: advancing a physically plausible prognostic state, and emitting the distributions that downstream users act upon. In practice, this means that the same network parameters $\theta$ are trained under two coupled (but conceptually distinct) signals: scores on $X_t$ that stabilize the transition operator, and scores on $Y_t$ that align the system with decision-relevant utilities. The composite loss in Section~\ref{sec:training} makes this separation explicit by assigning proper scoring rules to each component, and later experiments in Section~\ref{sec:exps} test whether this joint training indeed improves diagnostics without degrading the quality of the underlying dynamics. If diagnostics are included in the autoregressive state, they add little information about the physical dynamics while simultaneously increasing the dimension of the transition space and increasing estimator variance. 

Therefore, we are able to learn a function that simultaneously predicts the trajectory for further evolution ($X_t$) and required diagnostics for usability ($Y_t$) in a single pass. To make this problem tractable, we operate under a standard Markovian regime.

\textbf{Assumption 1 (First-order Markov property)}. We assume a 1st order Markov structure for the learning objective:

\begin{equation}
q_{\theta}\!\big(X_{1:T}, Y_{1:T}\mid X_{\le 0}, C\big)
= \prod_{t=1}^{T}
q_{\theta}\!\big(X_t, Y_t \mid X_{t-1},\, C\big)
\end{equation}

This formulation is a joint-distribution formulation, whereby we explicitly model both variables via the same model which share the same hidden representations. 

\begin{figure}[h]
    \centering
    \begin{minipage}[t]{0.47\textwidth}
        \centering
        \includegraphics[width=\linewidth, trim={0pt 40pt 0 40pt}, clip]{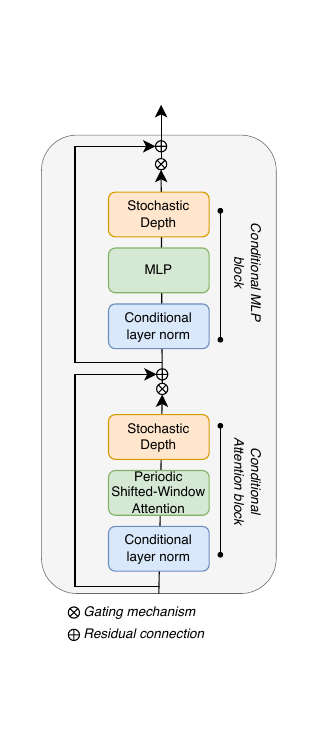}
        \caption{\textbf{Backbone of \method}. 
        Swin-style backbone with periodic shifted-window attention on a latitude--longitude grid.}
        \label{fig:swin_block}
    \end{minipage}
    \hfill
    \begin{minipage}[t]{0.47\textwidth}
        \centering
        \includegraphics[width=\linewidth, trim={0pt 40pt 0 40pt}, clip]{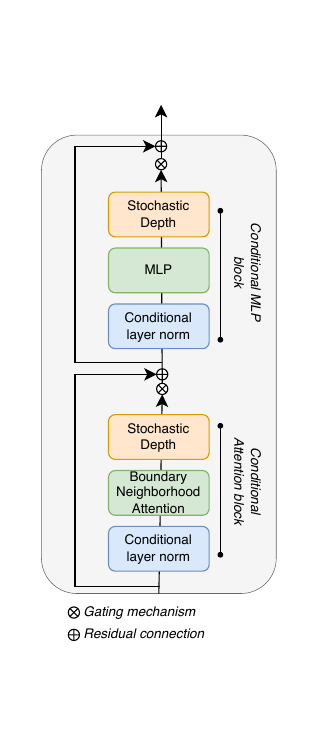}
        \caption{\textbf{Backbone of \methodtwo}. 
        Neighborhood Transformer with additional boundary conditions}
        \label{fig:other_block}
    \end{minipage}
    \vspace{-2mm}
    \rule{\linewidth}{.5pt}
\end{figure}

\subsection{Backbone of \method}

\begin{algorithm}[ht]
\caption{Periodic shifted-window attention}
\label{alg:periodic-swin}
\begin{algorithmic}[1]
\Require $x \in \mathbb{R}^{B \times C_\text{in{}} \times H_\text{grid} \times W_\text{grid}}$, window height $w_h$, width $w_w = 2w_h$, blocks $L$
\Ensure Updated tensor $x' \in \mathbb{R}^{B \times C_\text{in} \times H_\text{grid} \times W_\text{grid}}$
\State Partition $x$ into non-overlapping windows of size $(w_h, w_w)$ on the $(H_\text{grid},W_\text{grid})$ grid.
\For{$\ell = 1,\dots,L$}
    \State Set shift $s_\ell = (0,0)$ if $\ell$ even, else $s_\ell = (\Delta_h,\Delta_w)$.
    \State Apply shift $s_\ell$ with periodic wrap in longitude and zero-padding in latitude.
    \State Compute self-attention within each window with relative position bias.
    \State Mask attention entries that would couple northernmost and southernmost rows.
    \State Undo shift $s_\ell$ and merge window features back to the global grid.
\EndFor
\State $x' \gets x$ after the final merge.
\State \Return $x'$
\end{algorithmic}
\end{algorithm}

The backbone of \method is a Swin-style transformer \citep{liu2021swin} that produces a full-resolution forecast in a single pass, with many details originating from \citet{willard2025analyzing}. Given gridded inputs
$x \in \mathbb{R}^{B \times C_{\text{in}} \times H_\text{grid} \times W_\text{grid}}$ and a stochastic noise vector $z \sim \mathcal{N}(0, I)$, the network realizes a deterministic map
\[
(\hat{X}_t, \hat{Y}_t) \;=\; f_\theta(X_{t-1}, C, z).
\]

We write $(\hat{X}_t^{(n)}, \hat{Y}_t^{(n)}) \sim f_\theta(X_{t-1}, C)$ to denote that different draws of $z$ yield an ensemble of samples from the same parameterization. The sampling procedure is not explicitly conditioned on $\hat{X}_t$ to produce $\hat{Y}_t$; they are emitted jointly and we do not condition on the emission head on $X_t$ explicitly. The transformer operates in \emph{patch space}: inputs are embedded into a latent grid, processed by stacked windowed self-attention blocks with AdaLN-Zero conditioning, and then decoded back to the native resolution by a lightweight residual convolutional head. We use RoPE embeddings extended to 2D with periodic longitude positions to respect spherical geometry.

Uncertainty is represented by a low-dimensional noise vector $z$ injected through conditional layer norms, providing a parametric handle on aleatoric spread \citep{alet2025skillful}. Additional learned gates on residual connections modulate the effective depth and stabilize training.

Forecasting on a latitude--longitude grid imposes two simple geometric constraints that standard windowed attention does not enforce: (i) longitude is periodic; (ii) latitude has physical boundaries at the poles. To encode these constraints directly, while preserving the computational footprint of Swin, we introduce a variant of windowed self-attention, \emph{periodic shifted-window attention}.

A shallow decoder then maps the final patch embeddings back to the full-resolution grid, yielding joint probabilistic predictions for $X_t$ and $Y_t$ at each lead.

\subsection{Training objective and rollout scheme}
\label{sec:training}
The modeling choices in Section~\ref{sec:gem} are all geared toward a single goal: learn a probabilistic forecasting system whose conditional law over diagnostics $Y_t$ is useful for downstream decision-making.

\begin{greycustomblock}
\textbf{Decision-aligned Training Objective.}
We train $\theta$ by minimizing a composite loss that aggregates strictly proper scores over leads, variables, and spatial locations. For a given training sequence $(X_{\le 0}, C, X_{1:T}, Y_{1:T})$ and a rollout length $L \le T$, we unroll the model for $L$ steps under the Markov factorization, drawing an ensemble of size $S$ at each lead. The overall objective takes the form
\begin{equation}
  \mathcal{L}(\theta)
  =
  \mathbb{E}
  \left[
    \frac{1}{L}
    \sum_{t=1}^L
    \big(
      \mathcal{L}^{\text{prog}}_t
      +
      \mathcal{L}^{\text{diag}}_t
    \big)
  \right],
  \label{eq:training-objective}
\end{equation}
where the expectation is over forecast cases and model stochasticity. Each term $\mathcal{L}^{\text{prog}}_t$ penalizes errors in the prognostic fields $X_t$ and provides a learning signal for the transition operator; each term $\mathcal{L}^{\text{diag}}_t$ penalizes errors in the diagnostics $Y_t$ that enter decision rules. Both components use the same underlying strictly proper scoring rule, applied at different targets.
\end{greycustomblock}

For each training example and each lead $t$, we draw $S$ latent vectors $\{z^{(n)}\}_{n=1}^S$ and obtain an ensemble of samples. This ensemble is used to estimate proper scores for both the raw fields and any derived diagnostics.

\textbf{Representation of uncertainty}. For each input $(X_{t-1}, C)$ and lead $t$, we draw $S$ independent noise vectors $\{z^{(n)}\}_{n=1}^S$ and pass them through the stochastic transformer backbone (Section~\ref{sec:gem}) to obtain an ensemble $\{(\hat{X}_t^{(n)}, \hat{Y}_t^{(n)})\}_{n=1}^S$ that approximates the conditional law in Eq.~\eqref{eq:learning_objective}. We treat this ensemble as a Monte Carlo representation of $q_{\theta,t}(X_t, Y_t \mid X_{t-1}, C)$ and apply the fair CRPS estimator in Eq.~\eqref{eq:fcrps} within the composite objective in Eq.~\eqref{eq:training-objective}. penalizing both miscalibration and lack of sharpness for prognostic and diagnostic variables.

We employ two loss components to derive our final loss for both diagnostic and prognostic variables.

\vspace{1mm}
\paragraph{Loss component 1: Pixel-space marginal CRPS.}
Our primary score is the fair continuous ranked probability score (CRPS) applied to pixel-wise marginals. For a scalar target $x \in \mathbb{R}$ and ensemble samples $\{\hat{x}^{(n)}\}_{n=1}^S$, the unbiased estimator is
\begin{equation}
  \mathrm{fCRPS}\!\left(\{\hat{x}^{(n)}\}, x\right)
  =
  \frac{1}{S}\sum_{n=1}^S |\hat{x}^{(n)} - x|
  \;-\;
  \frac{1}{2S(S-1)}\sum_{n\neq n'} |\hat{x}^{(n)} - \hat{x}^{(n')}|.
  \label{eq:fcrps}
\end{equation}
This estimator is a strictly proper score for the predictive distribution induced by the ensemble: in expectation, it is uniquely minimized when the ensemble law matches the true conditional law of the target. We apply Eq ~\eqref{eq:fcrps} independently at each grid point, channel, and lead, with latitude-aware weighting to account for spherical geometry, and average the result to obtain $\mathcal{L}^{\text{prog}}_t$ and $\mathcal{L}^{\text{diag}}_t$.

\vspace{1mm}
\paragraph{Loss Component 2: Spectral log-power CRPS.}
Pixel-wise scores alone do not constrain the distribution of energy across spatial scales. To improve large-scale structure, we augment the pixel-space loss with a spectral term for all the variables. Given a field $X_t$ and its Real Spherical Harmonic Transform (RealSHT) coefficients $c_{\ell m}$, we define the degree-averaged power spectrum
\begin{equation}
  P_\ell(X_t)
  =
  \frac{1}{2\ell+1}
  \left(
    |c_{\ell 0}|^2
    +
    \sum_{m=1}^{\ell} 2|c_{\ell m}|^2
  \right),
\end{equation}
and apply fair CRPS in log-power space:
\begin{equation}
  \mathcal{L}_{\text{spec},t}
  =
  \sum_{\ell=1}^{L_{\max}}
  \frac{1}{\ell}
  \cdot
  \mathrm{fCRPS}
  \Big(
    \{\log P_\ell(\hat{X}_t^{(n)})\}_{n=1}^S,
    \log P_\ell(X_t)
  \Big).
\end{equation}

The $1/\ell$ weighting down-weights very small scales, and the logarithm balances contributions across orders of magnitude. We find that adding the spectral log-power CRPS helps to improve the power spectrum of the forecasts. We employ the spectral log-power CRPS as an additional loss component in \methodtwo.

\section{Validation}
\label{sec:exps}

Five aspects of our model deserve empirical investigation, and our goal in this section is to highlight them in turn:

\begin{enumerate}[leftmargin=2em, labelwidth=1em, labelsep=0.5em, itemsep=0pt, parsep=0pt]
    \item[\textbf{1.}] \textbf{Global skill.} We primarily care about decision-centric variables that users directly act upon. Does \method outperform operational NWPs and ML baselines? (Sec. \ref{subsec:prob_skill})
    
    \item[\textbf{2.}] \textbf{Stability.} Does \method maintain physical consistency and convergence to climatology on subseasonal timescales? (Sec. \ref{subsec:subseasonaltoseasonal})
    
    \item[\textbf{3.}] \textbf{Extremes and tail behavior.} Does the decision-aligned objective successfully capture heavy-tailed events without generative diffusion or multi-step fine-tuning strategies? (Sec. \ref{subsec:extreme_tails})
    
    \item[\textbf{4.}] \textbf{Utility.} Can a smaller model learn better variables for down-stream economic decision-making? (Sec. \ref{subsec:rev})
    
    \item[\textbf{5.}] \textbf{Case studies.} Can we demonstrate a real-world case study where a smaller and more lightweight model can match the detection of an event? (Sec. \ref{subsec:case_study})
\end{enumerate}

The validation setup therefore mirrors the structure of the decision problem in Section~\ref{sec:decision}. We specify derived variables and functionals that correspond directly to $Y_t$ and to simple transformations of $(X_t, Y_t)$, and we assess their predictive distributions using strictly proper scores. We focus on Tmax and Tmin as our primary variables of interest, since they best reflect the structural underestimation or overestimation discussed in Sec. \ref{sec:decision}. In addition, we include three other useful metrics for decision-making: Precipitation, Windspeed, and 500 hPa Geopotential, where the structural advantages are much less clear, to have as a baseline.

\textbf{Evaluation methodology}. 
Table~\ref{tab:validation_contract} maps each guiding question to a concrete diagnostic, metric, and experiment. The output of our weather model are samples from a joint predictive distribution over a high-dimensional atmospheric state (Eq. \ref{eq:learning_objective}). In practice, we are usually only interested in some lower-dimensional summary of the joint distribution. 
To evaluate the model's performance, we typically do the following:
\begin{enumerate}
  \item Specify the target random variable(s) or functional(s) that encode the question of interest (e.g., an exceedance event, a spatial average, a quantile).
  \item Obtain their predictive distribution by marginalizing or transforming the joint forecast, and assess it with appropriate proper scoring rules and diagnostic visualizations.
\end{enumerate}

Unless noted otherwise, all CRPS and Quantile Score skill values in Section~\ref{sec:exps} are computed relative to an ERA5-based climatological ensemble constructed to mirror each experiment in Table~\ref{tab:validation_contract}. To construct a climatology baseline, we generate 200 ensemble members for each forecast initialization date by sampling historical initialization dates from ERA5 within a 50-year look-back window and $\pm 3$ days of the forecast calendar day. This historical distribution is sampled without replacement, with a weighting scheme to preferentially draw from recent years to account for non-stationarity. From each sampled date, the subsequent contiguous days of the historical record form a complete forecast trajectory, preserving temporal and multi-variate consistency. Consequently, the climatology curves in Figures~\ref{fig:crps_skill} and~\ref{fig:s2s_skill_gem_gefs} represent a lead- and season-aware baseline that shares the same targets and diagnostics as the decision-aligned forecasts in Eq.~\eqref{eq:learning_objective}. 

Inter-model comparisons were performed over two distinct periods. For the operational NWP models, we use daily forecasts from a two year period from 2023-06-30 to 2025-08-3, which marks the transition to the 48r1 release cycle of the IFS when extended range forecast frequency was increased to daily. For the ML models, a much shorter 4-month period from 2025-07-01 to 2025-10-31, marking the start of the operational AIFS-ENS forecasts.

\begin{table*}[t]
\centering
\scriptsize
\renewcommand{\arraystretch}{1.2}
\begin{tabularx}{\textwidth}{
  >{\raggedright\arraybackslash}p{3.5cm}
  >{\raggedright\arraybackslash}p{3.0cm}
  >{\raggedright\arraybackslash}p{2.5cm}
  >{\raggedright\arraybackslash}p{1.6cm}
  >{\raggedright\arraybackslash}p{1.7cm}
}
\toprule
\textbf{Guiding question} 
& \textbf{Derived variable(s)} 
& \textbf{Metric(s)} 
& \textbf{Source of truth} 
& \textbf{Experiment(s)} \\
\midrule

How accurate are forecasts globally?
& Daily Tmax, Tmin, precip, 10m wind, Z500 (global, land, NA stations)
& CRPS, CRPS skill vs climatology
& ERA5, surface stations
& Fig.~\ref{fig:crps_skill}, Fig.~\ref{fig:station_absolute_scores_crps} \\

\addlinespace
How well do we forecast extremes?
& Tail diagnostics: Q95 (Tmax, precip, wind, Z500), Q05 (Tmin)
& Quantile Score (QS) skill at fixed quantiles
& ERA5
& Fig.~\ref{fig:extreme_skill} \\

\addlinespace
How skillful are the forecasts on sub-seasonal and seasonal timescales?
& 1–28d rolling means of core diagnostics
& CRPS skill vs climatology (per averaging window)
& ERA5
& Fig.~\ref{fig:s2s_skill_gem_gefs} \\

\addlinespace
How well do we capture high-impact events?
& Exceedance indicators (e.g.\ 98/99th pct wind or 3d precip) over event regions
& Probabilistic event maps; visual overlap with extremes
& ERA5
& Figs.~\ref{fig:beryl_forecast}, \ref{fig:melissa_forecast}, \ref{fig:european_floods} \\

\addlinespace
Are samples physically consistent across scales?
& Power spectra of Z500, Tmax, precip
& Spectral CRPS / spectral match vs ERA5
& ERA5
& Fig.~\ref{fig:power_spectra} \\

\addlinespace
How do we compare to baselines?
& Same diagnostics as above (global/land domains)
& CRPS / QS differences vs NWP and ML systems
& ERA5, stations
& Figs.~\ref{fig:crps_skill}, \ref{fig:scores_ai} \\

\addlinespace
Which design choices matter?
& Same diagnostics under variant architectures / training setups
& $\Delta$CRPS vs base configuration
& ERA5
& Figs.~\ref{fig:conditioning_vars}--\ref{fig:healpix}, Tab.~\ref{tab:ablations_conditional} \\

\bottomrule
\end{tabularx}
\caption{\textbf{Succinct validation map summary.} Each row maps a decision-aligned question to the derived variables, metrics, data source, and concrete experiment(s) in the paper.}
\label{tab:validation_contract}
\end{table*}

\subsection{Global performance on decision-relevant skill}
\label{subsec:prob_skill}

We first ask whether \method provides competitive global skill when judged as a probabilistic ensemble.

\begin{figure}
    \centering
    \includegraphics[width=\linewidth]{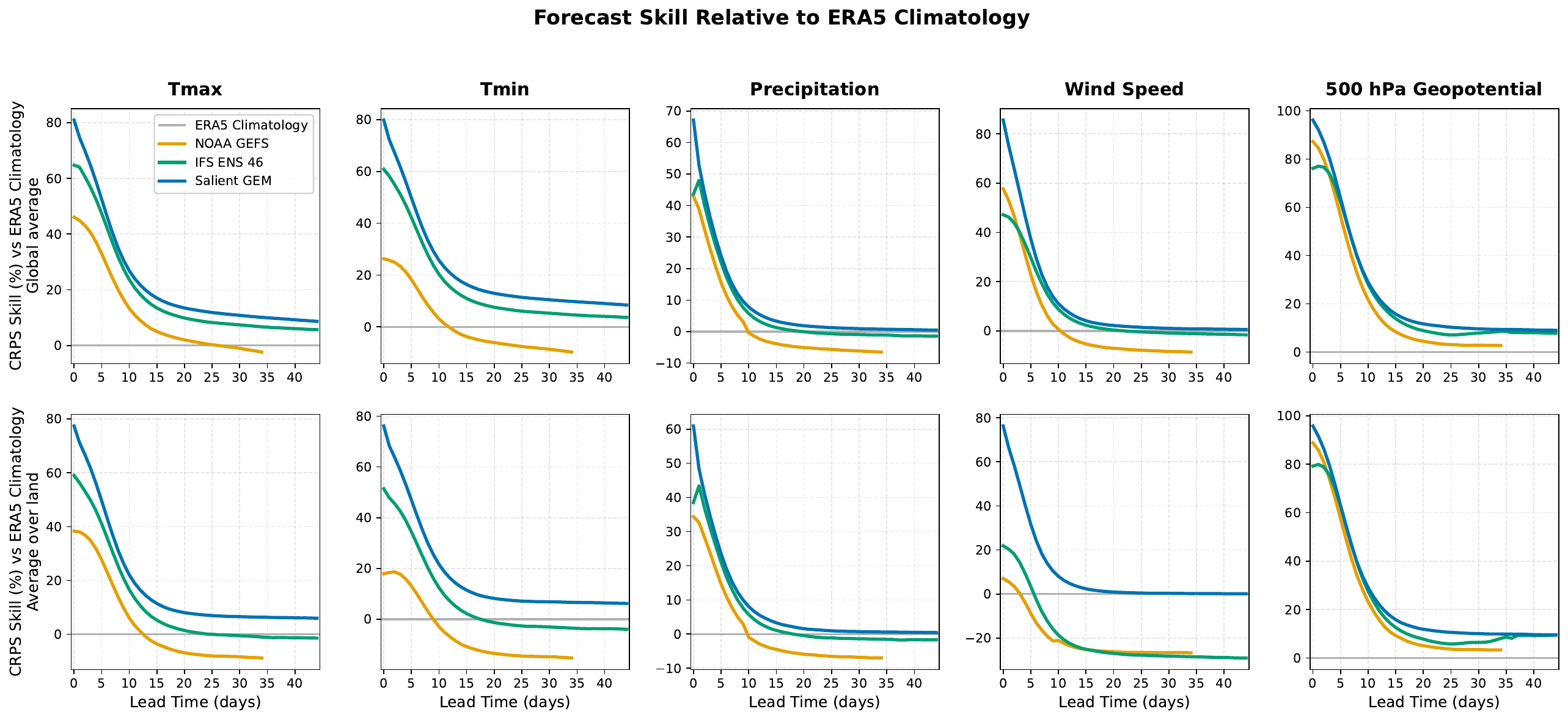}
    \caption{
    \textbf{CRPS skill scores relative to climatology for \method and operational NWP models} (NOAA GEFS, IFS ENS 46) across five variables: maximum temperature (Tmax), minimum temperature (Tmin), precipitation, 10m wind speed, and 500 hPa geopotential height. Positive values indicate improvement over climatological forecasts. Scores are latitude-weighted and averaged over 2023-06-30 to 2025-08-31. Top row: global average; bottom row: land-only average. \method maintains substantial skill advantage over NWP baselines across all variables and lead times.}
    \label{fig:crps_skill}
\end{figure}

Figure~\ref{fig:crps_skill} reports continuous ranked probability score (CRPS) skill relative to climatology for daily maximum and minimum 2m temperature, daily-accumulated precipitation, 10m wind speed, and 500\,hPa geopotential height, comparing \method to NOAA GEFS and IFS ENS 46.
Globally and over land, \method achieves higher CRPS skill than both NWP baselines across all variables and lead times out to 40\,days. Skill decays smoothly with lead time and approaches climatology, with no early collapse or late-time oscillations.

To decouple performance from potential reanalysis-specific artifacts, we also verify against global GHCN station observations over (Figure~\ref{fig:station_absolute_scores_crps} in Appendix~\ref{sec:station_validation} ).
Against stations, \method achieves the lowest absolute CRPS for all surface variables and lead times, with especially strong improvements for daily Tmax and Tmin.
Verification against ERA5 over land grid points (bottom row) yields a consistent ranking, suggesting that gains are not an artifact of the reanalysis training target.

\begin{figure}
    \centering
    \includegraphics[width=1\linewidth]{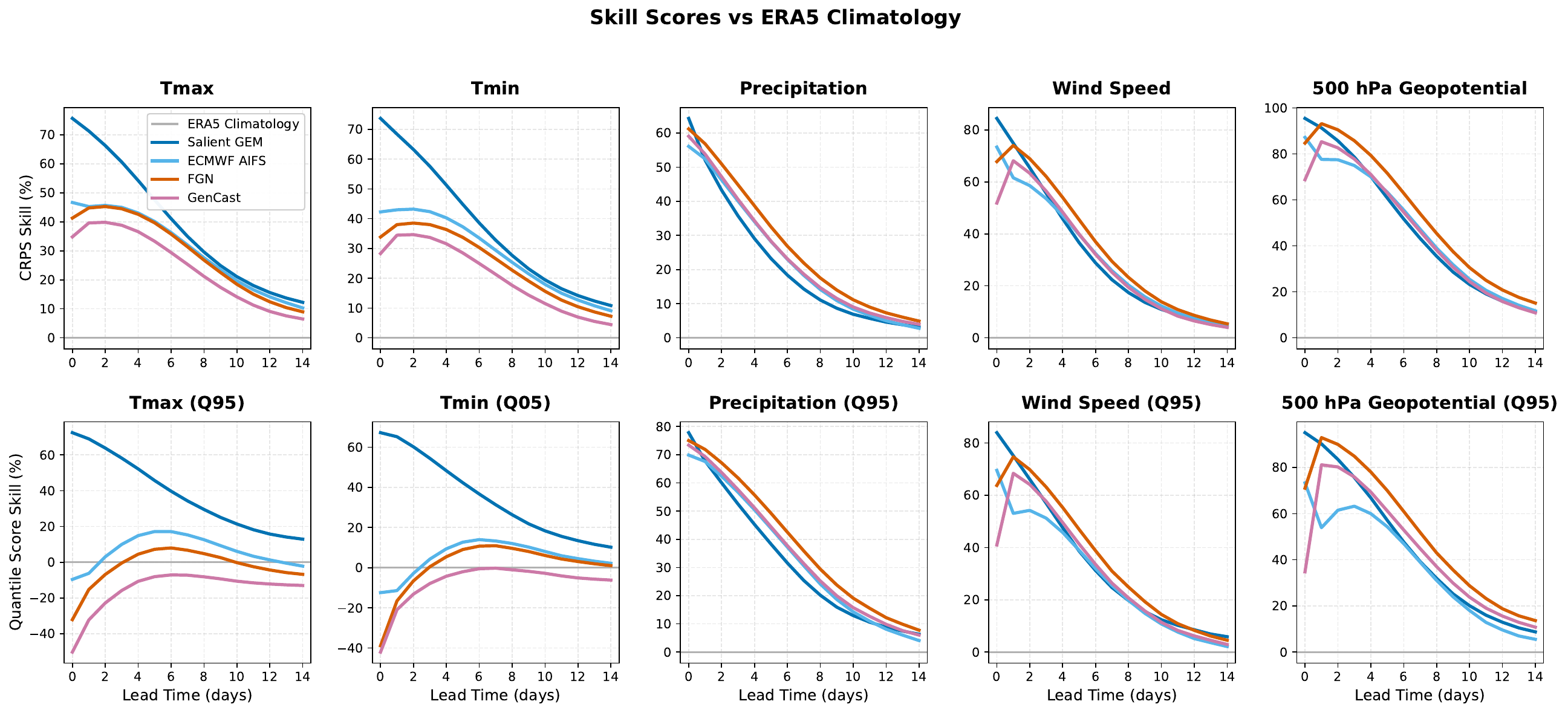}
    \caption{\textbf{CRPS skill (top) and Quantile Score skill at extreme quantiles (bottom) relative to climatology, comparing \method to similar ML models.} Q95 is used for Tmax, precipitation, wind speed, and geopotential (right-tail extremes); Q05 for Tmin (left-tail extremes).
    ML models that forecast only temperature snapshots are supplemented with tmin/tmax estimates derived from ERA5 climatological diurnal range (see Fig.~\ref{fig:tmin_tmax_deviation}; sub-daily outputs are aggregated to daily means via trapezoidal quadrature. Scores are latitude-weighted and averaged over a period from 2025-07-01 to 2025-10-31.} 
    \label{fig:scores_ai}
\end{figure}

\subsection{Forecasting extremes and tail behavior}
\label{subsec:extreme_tails}
To test whether \method captures the tails of the distribution, we evaluate Quantile Score (QS) skill at fixed extreme quantiles: Q95 for Tmax, precipitation, wind speed, and geopotential (right tails) and Q05 for Tmin (left tails).

\begin{figure}
    \centering
    \includegraphics[width=\linewidth]{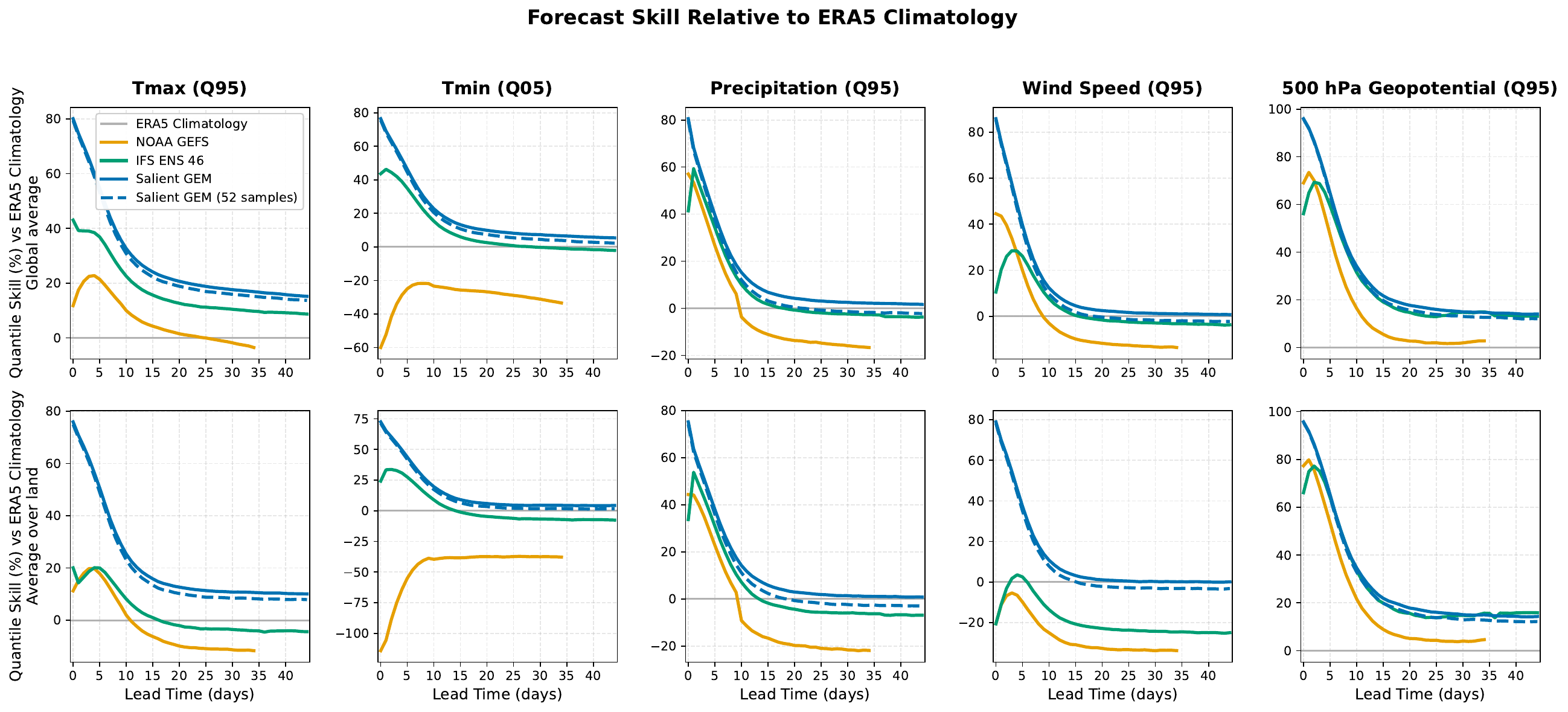}
    \caption{
    \textbf{Quantile Score skill relative to climatology}. Q95 is used for Tmax, precipitation, wind speed, and geopotential (right-tail extremes); Q05 for Tmin (left-tail extremes). Scores are latitude-weighted and averaged over a period from 2023-06-30 to 2025-08-31. Top row: global; bottom row: land-only.  
    }
    \label{fig:extreme_skill}
\end{figure}

Figure~\ref{fig:extreme_skill} shows QS skill relative to climatology.
Across both global and land-only domains, \method maintains positive skill at these tail quantiles for all variables and lead times up to 40\,days, again outperforming GEFS and IFS. We also compare two ensemble sizes (full ensemble with 200 samples versus 52 samples); the curves are nearly indistinguishable, indicating that tail skill is driven primarily by the learned conditional distribution rather than by lower CRPS due to larger brute-force sampling.

\paragraph{Note on methodology} All ML and NWP systems in Figures~\ref{fig:crps_skill}, \ref{fig:extreme_skill}, and~\ref{fig:scores_ai} are scored under a common verification setup: forecasts are interpolated to the \method\ $0.25^\circ$ latitude--longitude grid, with a shared land--sea mask, lead times, and diagnostic definitions as summarized in Table~\ref{tab:validation_contract}. For probabilistic ML baselines that do not natively emit daily extrema (e.g., GenCast, FGN, FourCastNet-3 in Table~\ref{tab:prob_models_compact_cited}), we reconstruct $T_{\max}$ and $T_{\min}$ from mean-temperature fields using ERA5 climatological diurnal ranges (Figure~\ref{fig:tmin_tmax_deviation}), which preserves mean skill but may underestimate tail sharpness those systems could attain with native extrema outputs. Daily cumulative precipitation, average wind speed and average 500 HPa geopotential are obtained by aggregating sub-daily outputs via trapezoidal quadrature. NWP ensembles (GEFS, IFS ENS 46) retain their operational configurations; when matching ensemble sizes (e.g., the 52-member subset in Figure~\ref{fig:extreme_skill}), we subsample \method\ rather than upsample baselines, so reported gaps slightly favour the reference systems. All cross-model skill differences should therefore be interpreted as performance under a harmonized verification pipeline rather than as a fully configuration-matched comparison.

\begin{customblockquote} 
\paragraph{Takeaway.} We observe that our performance gains are most pronounced precisely where state-of-the-art baselines suffer from structural bias: the estimation of daily extrema (Tmax and Tmin). Existing snaps-shot based methods systematically misestimate these quantities due to sparse temporal sampling that we avoid by structurally modeling them as a part of the model.
\end{customblockquote}

\subsection{Subseasonal-to-seasonal skill}
\label{subsec:subseasonaltoseasonal}

Because of the chaotic nature of the atmosphere, all forecasts lose skill with lead time and eventually become indistinguishable from a simple climatological baseline. To assess how \method behaves on S2S timescales (2–5 weeks) and beyond, we follow \citet{buizza2015}, who evaluate IFS ENS at long leads by looking at convergence of CRPS towards a climatological ensemble (\ref{sec:exps}).
They define the \emph{forecast-skill horizon} as the lead at which a bias-corrected ensemble no longer achieves significantly lower CRPS than climatology. They analyse 32-day lead forecasts over a 1-year period, using a 20-year weekly reforecast set (1992–2011) to estimate model bias and climatology. With temporal averaging windows up to 16 days and coarser spatial scales, the forecast-skill horizon for large-scale upper-air fields such as Z500 increases from about 2–3 weeks for daily grid-point fields to about 3–4 weeks.

\begin{figure}
    \centering
    \includegraphics[width=1\linewidth]{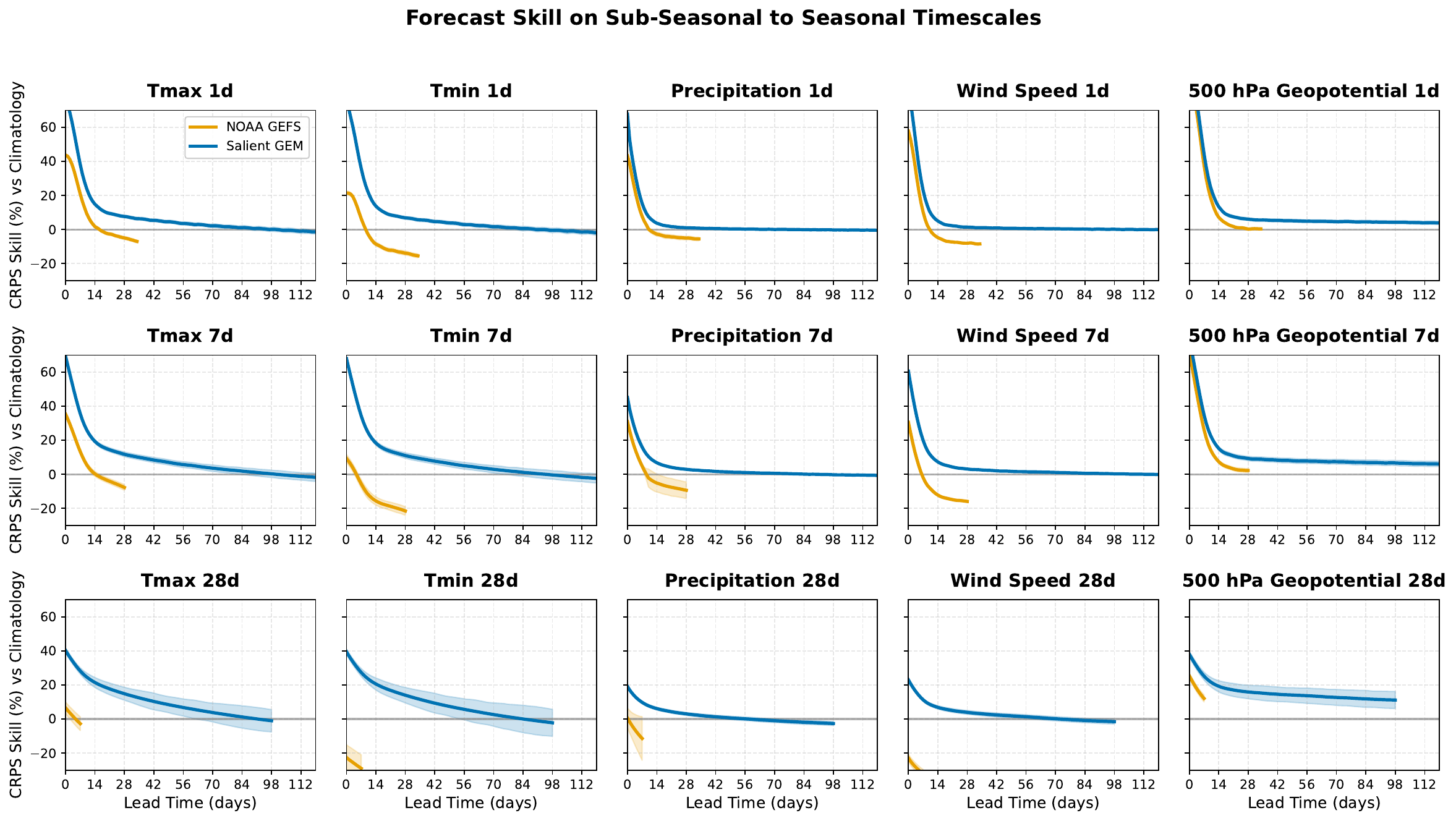}
    \caption{\textbf{\method and NOAA GEFS CRPS skill scores on S2S and seasonal timescales computed relative to climatology for variables averaged across temporal windows of increasing length (daily to monthly).} Scores are averaged over a 5 year period (2020-10-17 to 2025-06-30). The uncertainty bands correspond to the standard error for the mean, adjusted for lag 1 autocorrelation. Longer averaging windows reveal persistent skill at extended leads: \method maintains positive skill
   even weeks to months into the future, depending on the variable and the length of the temporal window while GEFS is often substantially worse than climatology.} \label{fig:s2s_skill_gem_gefs}
    
\end{figure}

We run an analogous experiment with \method using our climatological baseline as reference, but differ from \citet{buizza2015} in that we (i) focus on decision-relevant surface fields (plus Z500), (ii) only use temporally averaged fields at fixed $0.25^\circ$ resolution, and (iii) evaluate nearly five years of forecasts (Oct 2020 to June 2025) out to the full \method lead range, with 200 samples for both forecast and climatological ensembles. Figure~\ref{fig:s2s_skill_gem_gefs} shows CRPS skill for \method and NOAA GEFS, with rows corresponding to different averaging windows; the rolling means use a 1-day stride, so at lead $T$ the value reflects the score for the average from $T$ to $T$ plus the window length.

There are a few salient aspects of the plot. First, in every case we see two distinct regimes in the skill-score curves: a \emph{fast-decay regime} up to about 10–15 days and a \emph{slow-decay regime} thereafter. This structure reflects the fact that the atmosphere can be viewed as the sum of a fast component (synoptic and convective systems) and a slow component (large-scale, low-frequency anomalies such as SST patterns, soil moisture/snow, quasi-stationary waves, and intraseasonal tropical variability).

In the fast-decay regime, error growth is dominated by the synoptic flow: small initial-condition perturbations grow with e-folding times of a few days, so CRPS skill drops rapidly and the ensemble quickly spans the climatological spread of fronts, storms, and jet paths. After ~10 days, members are already sampling essentially all plausible synoptic realizations compatible with the slow background state. Beyond this point, the remaining skill comes from the slow components, whose evolution is much less chaotic and decays on timescales of  weeks-to-months; correspondingly, the skill curves show a much shallower decline as low-frequency anomalies gradually relax toward climatology.

After the kink, normalized skill for \method remains positive well into the subseasonal range. For near-surface temperature, the crossover to zero occurs only near the end of our forecast range, whereas noisier fields such as precipitation and wind have shorter horizons: roughly 30–40 days for daily means, increasing to about 40–50 days for 7-day precipitation and 50–60 days for 7-day wind-speed averages; for Z500, the slow-decay tail does not cross zero within the plotted range. These forecast-skill horizons are sensitive to the definition and calibration of the climatological ensemble, and we have not optimized the baseline here, so the numbers should be read as indicative rather than sharp limits; nonetheless, they are broadly consistent with \citet{buizza2015}, who find 2–3-week horizons for instantaneous upper-air fields extending toward a month with temporal and spatial averaging, despite differences in the variables we look at and in how we construct the climatological ensemble.

In comparison to \method, GEFS typically loses skill relative to climatology after only 1–2 weeks for surface fields, and for coarsened minimum temperature, precipitation, and surface wind the skill is often negative even at one-week lead, indicating substantially poorer calibration and weaker exploitation of low-frequency predictability than \method.

\textbf{Sources of S2S skill.}
GEM-2 uses a relatively minimal set of traditional S2S input predictors: a single SST field plus stratospheric level data up to 10 hPa, both of which are simply evolved with the prognostic fields at each timestep. We find that with this trivial coupling scheme, the SST field is prone to accumulation of mesoscale energy in the western boundary currents and Antarctic circumpolar current. We handle this by instead predicting a spatially low-passed version of the SST field, which still gives the model context about the evolution of the SST state from its initial conditions, and allows for a limited form of atmosphere to ocean feedback. We are exploring more sophisticated coupling mechanisms and additional ocean and land predictors for future versions of the model.

\textbf{Why convergence to climatology matters.}
 Existing NWP ensembles routinely perform \emph{substantially worse} than climatology at longer lead times (as seen in Figure~\ref{fig:s2s_skill_gem_gefs}), with forecast distributions that continue to deviate from climatology long after their marginal skill has vanished. For part of the horizon they are therefore systematically inferior to a trivial reference, making it unclear when and how downstream users should trust the system. By contrast, when a system at each lead up to some horizon obtains forecasts that are at least as good as a climatology baseline, users can rely on a single probabilistic system over that range without having to splice in a separate baseline or risk systematic underperformance relative to a simple reference.

\begin{customblockquote} 
\paragraph{Takeaway.} Contrary to operational ensembles that frequently degrade below climatological baselines at subseasonal scales, we find that GEM-2 exploits low-frequency predictability to maintain positive skill weeks into the future , enabling a single probabilistic system that stably converges to the climate mean.
\end{customblockquote}

\subsection{Relative Economic Value}
\label{subsec:rev}

A common way to represent the potential value of a forecast is to compute the relative economic value for different cost/loss ratios \citep{richardson2000skill}. Briefly, REV measures the benefit of using a particular forecast over using a simple climatology baseline. A REV of 0 implies that the forecast is no better than the climatology baseline, and a REV of 100\% means a perfect forecast. The different cost/loss ratios can show the trade-offs decision-makers use by showing the impact of a particular forecast across many different decision scenarios. The \textit{cost} is the cost of taking preventative action. The \textit{loss} is the financial loss incurred if the weather event happens. Clearly, a rational user is willing to pay some cost to prevent a financial loss, but it cannot be larger than the loss itself. Therefore, the cost/loss ranges from small ratios (approx $10^{-3}$) to $10^0$, implying different thresholds for risks. A user should take action, on the basis of expected utility, if the predicted probability of an event happening is higher than the cost/loss ratio. 

To compute the REV metric, we follow the relative economic value calculation of \citet{price2023gencast}. Briefly, we define the event as a binary event based on the variables and quantiles described in Fig. \ref{fig:skill_scores_REV}. We obtain an ensemble of events and use these ensembles to compute a predicted probability of the event occurring based on the predicted probability. Then, we use each cost/loss ratio as a decision threshold, and plot a variety of cost/loss ratios and REVs. We evaluate REV against two models: FGN and IFS ENS46. We do so to primarily declutter the number of models in the figure, since FGN is a successor to GenCast and has superior CRPS values. The results are presented in Fig. \ref{fig:skill_scores_REV}. 

\paragraph{Interpretation} A general trend is visible across most forecasts: the values degrade as lead time increases. This suggest that all models provide less economic value at longer horizons due to increased uncertainty in the predictions. For precipitation, windspeed, and Z500 quantiles, the models struggle to provide value for events with a cost/loss of $< 10^{-2}$. It is useful to think of these events as high-stakes decisions where users are extremely risk-averse and the probability of an event happening must be quite small for users to take action. Similarly, for most variables, the REV metric drops for events near $10^0$. It's useful to think of such decisions as being low-risk events, where a very high probability of an event must be predicted for the user to take action. \method performs best in the short-range temperature forecast as well as for most leads for Tmax quantiles. FGN is superior at longer lead times (leads 7 and 14), especially for Z500, windspeed, and precip. Despite the smaller model size and simpler and lightweight training objective, we obtain competitive performance to FGN and outperform IFS ENS for most variables.

\begin{customblockquote} 
\paragraph{Takeaway.} Contrary to the assumption that maximizing downstream value requires computationally intensive generative priors, we find that our lightweight, decision-aligned framework outperforms operational ensembles and remains competitive with larger, more expensive models. This suggests that parsimonious transformers can be sufficient to generate state-of-the-art economic value.
\end{customblockquote}

\subsection{Event-scale performance and case studies}
\label{subsec:case_study}

We further examine the evolution of forecast probability maps for several major events, using event-specific diagnostics derived from the joint state.

\begin{figure}
    \centering
    \includegraphics[width=1\linewidth]{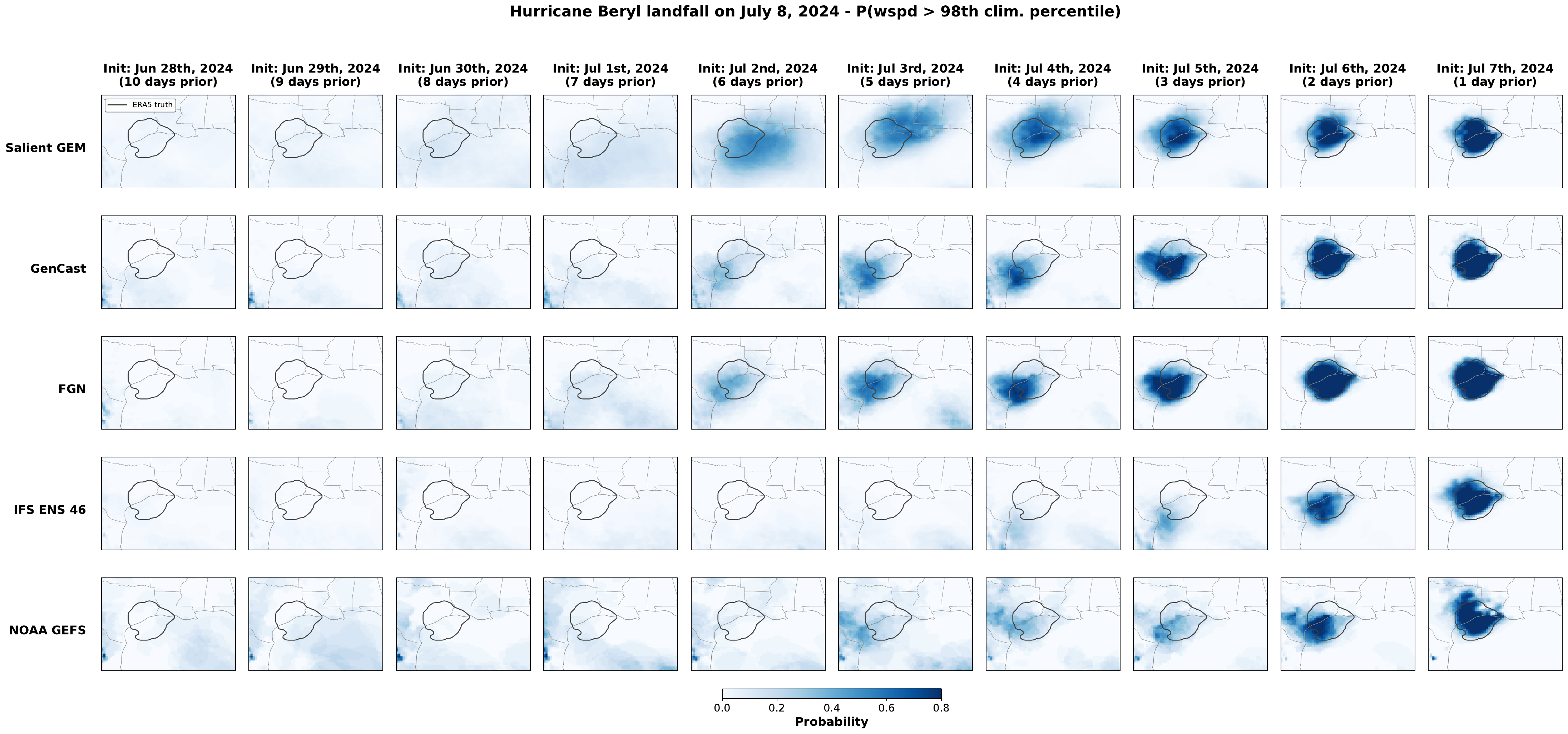}
    \caption{\textbf{Forecast evolution for Hurricane Beryl's Texas landfall (July 8, 2024), showing probability of exceeding 98th percentile wind speed.} Black contour: ERA5-observed extreme wind region on July 8th. \method detects the landfall location 5–6 days in advance with smoothly evolving probabilities, while IFS ENS 46  and NOAA GEFS do not localize the event until 2 days prior.} 
    \label{fig:beryl_forecast}
\end{figure}

\begin{figure}
    \centering
    \includegraphics[width=1\linewidth]{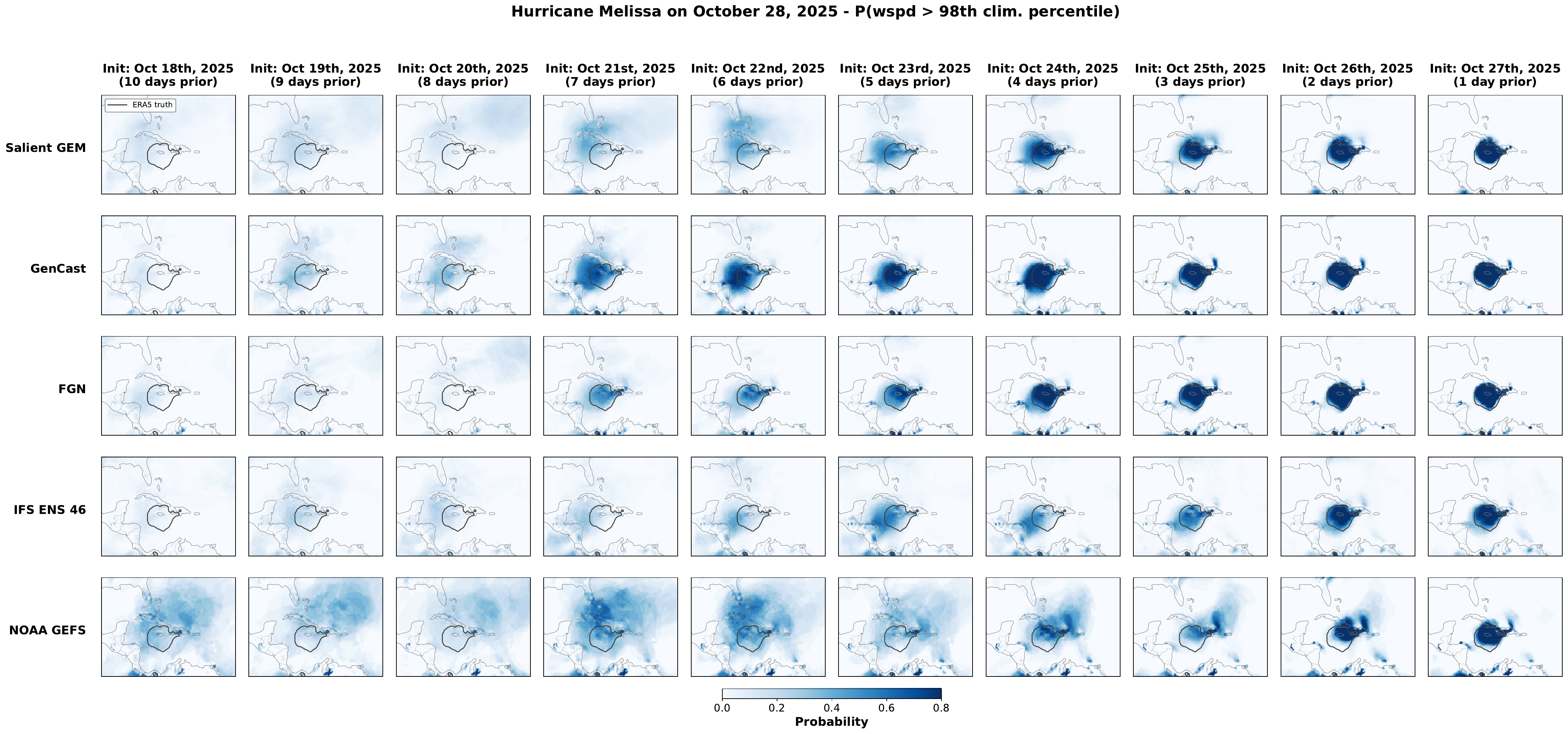}
    \caption{\textbf{Forecast evolution for Hurricane Melissa (October 28, 2025), showing probability of exceeding 98th percentile wind speed from 10 to 1 day prior.} Black contour: ERA5-observed extreme wind region. \method produces focused, high-probability forecasts beginning 7–8 days in advance, tracking the storm's path through the Gulf of Mexico. IFS ENS 46 shows similar localization but with slightly more spread at longer leads. NOAA GEFS exhibits diffuse probability across the entire basin at extended leads, with sharpening only at short range.}
    \label{fig:melissa_forecast}
\end{figure}

\paragraph{Tropical cyclones.}
For tropical cyclones we consider the probability that daily-maximum 10m wind speed exceeds its local 98th percentile, computed relative to a historical ERA5 climatology.
Figures~\ref{fig:beryl_forecast} and~\ref{fig:melissa_forecast} display this probability field for Hurricane Beryl (Texas landfall, July 8, 2024) and Hurricane Melissa (October 28, 2025).

For Beryl (Figure~\ref{fig:beryl_forecast}), \method begins to localize the landfall region along the Texas coast 5–6 days in advance, with a compact lobe of elevated probability that sharpens smoothly over time.
In contrast, IFS ENS 46 and GEFS remain diffuse over the Gulf of Mexico until roughly two days before landfall, at which point they rapidly re-focus.
Melissa (Figure~\ref{fig:melissa_forecast}) shows a similar pattern: \method tracks the storm’s path with concentrated probability from 7–8 days ahead, IFS exhibits a comparable but slightly broader signal, and GEFS maintains a basin-wide, low-amplitude pattern until short range.
These behaviors mirror the global tail scores: \method allocates probability mass earlier and more selectively over plausible tracks.

\begin{figure}
    \centering
    \includegraphics[width=1\linewidth]{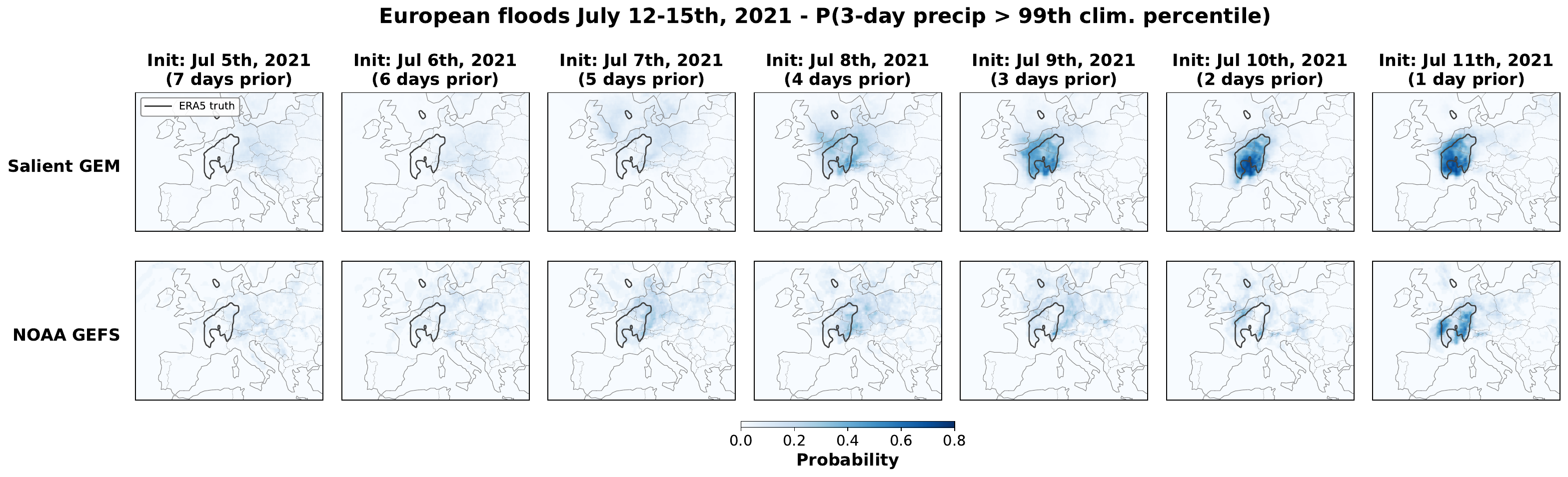}
    \caption{\textbf{Forecast evolution for the July 2021 European floods (July 12–15), showing probability of exceeding 99th percentile 3-day precipitation.} Black contour: ERA5-observed extreme precipitation region over Germany, Belgium, and the Netherlands. \method localizes the event 4–5 days in advance with high confidence while NOAA GEFS shows diffuse probability across broader Europe without comparable spatial precision.}
    \label{fig:european_floods}
\end{figure}

\paragraph{Heavy precipitation and flooding.}
For heavy precipitation we evaluate the probability that 3-day accumulated precipitation exceeds its local 99th percentile.
Figure~\ref{fig:european_floods} shows forecasts for the July 12–15, 2021 European floods.
Four to five days prior, \method concentrates probability over western Germany, Belgium, and the Netherlands, overlapping the ERA5-observed extreme region.
GEFS, by contrast, maintains a broad swath of moderate probability spanning much of Europe, with less spatial specificity.

›

\begin{figure}
    \centering
    \includegraphics[width=1\linewidth]{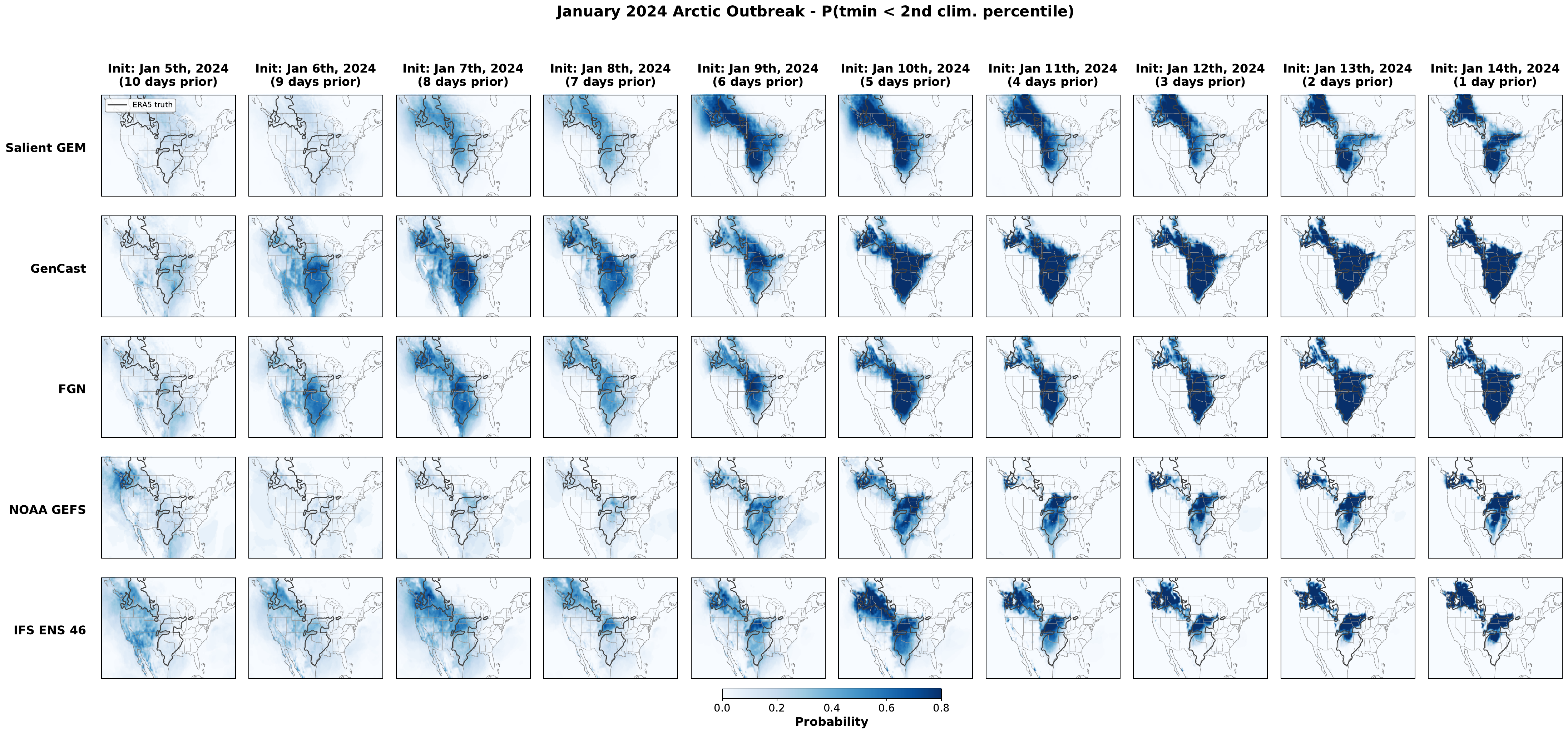}
    \caption{
\textbf{Forecast evolution for the January 2024 Arctic outbreak}: probability of Tmin below the 2nd climatological percentile (colder than 98\% of historical days). Black contour: ERA5-observed extreme cold region. \method produces spatially precise, temporally consistent forecasts 7–8 days in advance. NOAA GEFS shows localized signals but lacks temporal consistency across initializations. IFS ENS 46 approaches GEM's performance but underestimates the full spatial extent of the extreme cold region.
    }
    \label{fig:arctic_outbreak}
\end{figure}

\textbf{Power Spectra}. Furthermore, we ask whether the samples produced by \method have sensible spatial structure across scales. Figure~\ref{fig:power_spectra} shows spherical power spectra for 500\,hPa geopotential height, Tmax, and precipitation at lead times of 1, 5, and 10 days, comparing ERA5 (black) with \method, IFS ENS 46, GenCast, and FGN.

\section{Related Work}
\label{sec:related_work}

Modern global weather forecasting sits at the intersection of two well-established paradigms. On the one hand, numerical weather prediction solves discretized physical equations with explicit time-stepping and carefully engineered numerics; it is latency-bounded, traceable, and widely operationally employed, but is increasingly outpaced on pure skill by \textit{learned} weather emulators. On the other hand, recent machine learning systems learn forecast operators directly from reanalysis and observations. They are learned in both a deterministic form \citep{pathak2022fourcastnet, bi2023pangu, chen2023fuxi, chen2025fengwu} or as probabilistic generators \citep{price2024probabilistic, alet2025skillful, bonev2025fourcastnet3, couairon2024archesweather}.

\textbf{Deterministic ML Weather Models:} Recent years have seen a surge of deterministic data-driven global forecasters that rival numerical weather prediction (NWP) in accuracy. \textit{FourCastNet} introduced high-resolution global forecasting with Fourier neural operators, demonstrating competitive skill up to 10 days \citep{pathak2022fourcastnet}. \textit{Pangu-Weather} employed 3D convolutional networks to achieve state-of-the-art medium-range forecasts (0–10 days) on 0.25° grids \citep{bi2023pangu}. Graph neural network approaches have also emerged: an early GNN prototype by \citet{keisler2022gnn} explored global forecasting on irregular meshes, and \textit{GraphCast} later achieved excellent 10-day accuracy using a learned graph message-passing over the sphere \citep{lam2023learning}. More recently, \textit{FengWu} leveraged a multi-modal transformer with replay buffer training to push deterministic skill slightly beyond 10 days \citep{chen2025fengwu}, outperforming GraphCast and Pangu on most variables \citep{chen2025fengwu}. These deterministic models focus on raw forecast accuracy, but by design they produce a single trajectory and often require external ensembling or post-processing for uncertainty quantification.

\textbf{Probabilistic and Ensemble ML Models:} To represent forecast uncertainty, several probabilistic ML models generate ensembles of weather trajectories. \textit{GenCast} \citep{price2024probabilistic} introduced a diffusion-based 15-day ensemble model at 0.25° that surpassed the ECMWF 51-member ensemble (ENS) in skill while delivering 80+ variables in minutes. \textit{FGN} took a different approach, training an ensemble of perturbation-based models directly on the continuous ranked probability score (CRPS) \citep{alet2025skillful}. This method achieved state-of-the-art probabilistic skill and calibration, even matching human forecasters on tropical cyclone tracks \citep{alet2025skillful}. ECMWF’s \textit{AIFS-CRPS} (AI Forecasting System) similarly uses CRPS as a loss function with multiple ensemble members \citep{lang2024aifs}. \textit{FourCastNet 3} extended the original FourCastNet into a probabilistic regime with a geometric ensemble method, attaining improved calibration and stable spectra up to 60 days \citep{bonev2025fourcastnet3}. Similarly, \textit{ArchesWeatherGen} combined a deterministic core (ArchesWeather) with a flow-matching generative model to sample realistic 15-day outcomes, outperforming the ECMWF ENS at 1.5° resolution \citep{couairon2024archesweather}. Other work has explored diffusion and generative adversarial techniques for ensemble forecasting – for example, \citet{li2024generative} used diffusion to emulate ensemble spread from a control forecast, and \textit{FengWu}'s developers built a diffusion-based ensemble (\textit{FengWu-ENS}) conditioned on their deterministic model to improve reliability \citep{chen2025fengwu}. Meanwhile, \textit{FuXi-Ens} extended the FuXi system to probabilistic forecasts via an ensemble of ML models \citep{zhong2024fuxiens}, also showing benefits over traditional ensembles. Furthermore, some models have demonstrated stable 6-hourly forecasts out to 75 days, an example of recent progress in extending S2S horizons \citep{stock2025swift} or forecasting 44 days ahead single-shot \citep{nguyen2025omnicast}.

\textbf{Hybrid and Foundation Approaches:} Some systems integrate learning with traditional pipelines or multi-task objectives. \textit{FuXi Weather} is an end-to-end system that combines machine learning with data assimilation: it cycles observational assimilation and ML forecasting, achieving 10-day forecasts at 0.25° that in some cases exceed ECMWF’s high-resolution model \citep{chen2023fuxi, sun2025fuxiweather}. By using satellite data directly, FuXi can improve skill in observation-sparse regions \citep{sun2025fuxiweather}. GraphDOP learns weather forecasts directly from observations \citep{alexe2024graphdop}. In a different vein, \textit{NeuralGCM} proposed by \citet{kochkov2024neuralgcm} integrates neural networks into a general circulation model framework, bridging weather and climate timescales. Lastly, \textit{ClimaX} introduced the idea of foundation models for weather and climate—a single transformer pretrained on diverse climate data and fine-tuned for forecasting tasks (among others), which achieved competitive performance on WeatherBench benchmarks \citep{nguyen2023climax}. While such approaches are not yet topping operational metrics, they highlight the trend of embedding domain physics and multi-task learning into ML weather models

\begin{table*}[ht]
\centering
\scriptsize
\renewcommand{\arraystretch}{1.2}
\setlength{\tabcolsep}{4pt}
\begin{tabularx}{\textwidth}{@{}l c c c c c c c@{}}
\toprule
\textbf{Model} &
\textbf{Pred. Inputs} &
\textbf{Pred. Outputs} &
\textbf{Obj.} &
\textbf{Reg.} &
\textbf{Step Gen.} &
\textbf{Train Hor.} &
\textbf{Repr. Space} \\
\midrule

GenCast \citep{price2023gencast} &
$X_{t-12}, X_{t}$ &
$X_{t+12}$ &
score/diff. &
-- &
diff. &
1-step &
grid $0.25^\circ$ \\

FGN \citep{alet2025skillful} &
$X_{t-6}, X_t$ &
$X_{t+6}$ &
CRPS &
-- &
1-shot &
1-step (+AR) &
mesh $\sim 0.25^\circ$ \\

FourCastNet-3 \citep{bonev2025fourcastnet3} &
$X_t$ &
$X_{t+6}$ &
CRPS &
Spectral &
1-shot &
multi-step &
sphere $0.25^\circ$ \\

AIFS-ENS/CRPS \citep{lang2024aifs} &
$X_{t-6}, X_t$ &
$X_{t+6}$ &
CRPS &
-- &
1-shot &
1-step &
grid $\sim 0.25^\circ$ \\

ArchesGen \citep{couairon2024archesweather} &
$X_{t-24}, X_t$ &
$X_{t+24}$ &
flow-match &
-- &
1-shot &
1-step &
grid $1.5^\circ$ \\

DLWP \citep{weyn2021subseasonal} &
$X_{t-6}, X_t$ &
$X_{t+6}$ &
MSE &
-- &
1-shot &
1-step &
cubed-sphere $1.4^\circ$ \\

\hline
\rowcolor{gray!10}
\method (this work) &
$X_t, C$ &
$X_{t+24}, Y_{t+24}$ &
CRPS &
-- &
1-shot &
1-step &
grid $\sim 0.25^\circ$ \\
\rowcolor{gray!10}
\methodtwo (this work) &
$X_t, C$ &
$X_{t+24}, Y_{t+24}$ &
CRPS &
Spectral &
1-shot &
1-step &
grid $\sim 1^\circ$ \\

\bottomrule
\end{tabularx}
\caption{\textbf{Probabilistic ML forecasters.} 
Notation: $X_t$ = prognostic state; $Y_t$ = diagnostic targets (e.g.\ daily aggregates); $C$ = conditioning/forcing fields; ``score/diff.'' = diffusion score matching objective; ``CRPS+spec.'' = CRPS with spectral regularization. ``Spectral Reg.'' indicates explicit frequency-space regularization. ``Step Gen.'': ``diff.'' = multi-step diffusion sampling per forecast step, ``1-shot'' = single forward pass per step. ``Train Hor.'' is the optimization horizon used in training (single-step vs multi-step).}
\label{tab:prob_models_compact_cited}
\end{table*}

\section{Discussion}
\label{sec:discussion}

In this work, we operationalize a mechanism to jointly learn global atmospheric dynamics and user-centric variables. This is done by separating prognostic variables from diagnostic targets. The empirical results in Section \ref{sec:exps} show that this shift in objective is sufficient to exceed operational ensembles on CRPS and quantile metrics, capture extremes, and maintain coherent long-lead behavior, all while keeping the architecture lightweight and single-shot. Relative to prior probabilistic systems (Table \ref{tab:prob_models_compact_cited}), the contribution is less a new mechanism and more a change in what is being modeled: decision-level distributions, emitted natively and calibrated directly.

The underlying reason this works is structural. Jointly predicting $(X_t, Y_t)$ forces the model to represent the uncertainty in diagnostics conditioned on physically plausible trajectories, while the spectral component of the loss constrains energy across scales. The ablations in Table \ref{tab:ablations_conditional} indicate that this design is extremely robust to many design choices with only a few variables impacting overall performance.

At the same time, several limitations follow from the formulation. The model inherits the constraints of first-order Markov structure, fixed reanalysis training, and an implicit utility class tied to proper scoring rules. Diagnostics are limited to a small set, and utilities remain indirect. Extending the approach to richer diagnostic spaces, assimilating observations online, or conditioning training objectives on explicit cost–loss structures are potential extensions of this line of work. Nonetheless, the results here demonstrate that a simple probabilistic transformer, trained on the right objects, can supply actionable uncertainty information without heavy diffusion sampling, large ensembles, or complex multi-stage pipelines.

\clearpage
\bibliographystyle{abbrvnat} % or plainnat/unsrtnat
\bibliography{library}        % matches your library.bib

@article{lam2023learning,
  title={Learning skillful medium-range global weather forecasting},
  author={Lam, Remi and Sanchez-Gonzalez, Alvaro and Willson, Matthew and Wirnsberger, Peter and Fortunato, Meire and Alet, Ferran and Ravuri, Suman and Ewalds, Timo and Eaton-Rosen, Zach and Hu, Weihua and others},
  journal={Science},
  volume={382},
  number={6677},
  pages={1416--1421},
  year={2023},
  publisher={American Association for the Advancement of Science}
}

@article{price2023gencast,
  title={Gencast: Diffusion-based ensemble forecasting for medium-range weather},
  author={Price, Ilan and Sanchez-Gonzalez, Alvaro and Alet, Ferran and Andersson, Tom R and El-Kadi, Andrew and Masters, Dominic and Ewalds, Timo and Stott, Jacklynn and Mohamed, Shakir and Battaglia, Peter and others},
  journal={arXiv preprint arXiv:2312.15796},
  year={2023}
}

@article{pathak2022fourcastnet,
  title={Fourcastnet: A global data-driven high-resolution weather model using adaptive fourier neural operators},
  author={Pathak, Jaideep and Subramanian, Shashank and Harrington, Peter and Raja, Sanjeev and Chattopadhyay, Ashesh and Mardani, Morteza and Kurth, Thorsten and Hall, David and Li, Zongyi and Azizzadenesheli, Kamyar and others},
  journal={arXiv preprint arXiv:2202.11214},
  year={2022}
}

@article{bi2023pangu,
  title={Accurate medium-range global weather forecasting with 3D neural networks},
  author={Bi, Kaifeng and Xie, Lingxi and Zhang, Hengheng and Chen, Xin and Gu, Xiaotao and Tian, Qi},
  journal={Nature},
  volume={619},
  pages={533--538},
  year={2023}
}

@article{price2024probabilistic,
  title={Probabilistic weather forecasting with machine learning},
  author={Price, Ilan and Sanchez-Gonzalez, Alvaro and Alet, Ferran and Andersson, Tom R. and El-Kadi, Andrew and Masters, Dominic and Ewalds, Timo and Stott, Jacklynn and Mohamed, Shakir and Battaglia, Peter and Lam, Remi and Willson, Matthew},
  journal={Nature},
  volume={637},
  pages={84--90},
  year={2024}
}

@article{alet2025skillful,
  title={Skillful joint probabilistic weather forecasting from marginals},
  author={Alet, Ferran and Price, Ilan and El-Kadi, Andrew and Masters, Dominic and Markou, Stratis and Andersson, Tom R. and Stott, Jacklynn and Lam, Remi and Willson, Matthew and Sanchez-Gonzalez, Alvaro and Battaglia, Peter},
  journal={arXiv preprint arXiv:2506.10772},
  year={2025}
}

@article{couairon2024archesweather,
  title={Archesweather \& archesweathergen: a deterministic and generative model for efficient ml weather forecasting},
  author={Couairon, Guillaume and Singh, Renu and Charantonis, Anastase and Lessig, Christian and Monteleoni, Claire},
  journal={arXiv preprint arXiv:2412.12971},
  year={2024}
}

@article{bonev2025fourcastnet3,
  title={FourCastNet 3: A geometric approach to probabilistic machine-learning weather forecasting at scale},
  author={Bonev, Boris and Kurth, Thorsten and Mahesh, Ankur and Bisson, Mauro and Kossaifi, Jean and Kashinath, Karthik and Anandkumar, Anima and Collins, William~D. and Pritchard, Michael~S. and Keller, Alexander},
  journal={arXiv preprint arXiv:2507.12144},
  year={2025}
}

@article{kochkov2024neuralgcm,
  title={Neural general circulation models for weather and climate},
  author={Kochkov, Dmitrii and Yuval, Janni and Langmore, Ian and Norgaard, Peter and Smith, Jamie and Mooers, Griffin and Kl{\"o}wer, Milan and Lottes, James and Rasp, Stephan and D{\"u}ben, Peter and Hatfield, Sam and Battaglia, Peter and Sanchez-Gonzalez, Alvaro and Willson, Matthew and Brenner, Michael~P. and Hoyer, Stephan},
  journal={Nature},
  volume={632},
  number={8027},
  pages={1060--1066},
  year={2024}
}

@inproceedings{nguyen2023climax,
  title        = {ClimaX: A foundation model for weather and climate},
  author       = {Nguyen, Tung and Gupta, Jayesh K. and Chen, Shengchao and Bi, Kuan and Xie, Weizhen and Bi, Ke and Li, Xueyan and Chen, Rose Yu and Hassanzadeh, Pedram and Song, Jialin and others},
  booktitle    = {Proceedings of the 40th International Conference on Machine Learning (ICML)},
  volume       = {202},
  pages        = {11502--11521},
  year         = {2023}
}

@article{weyn2021subseasonal,
  title={Sub-seasonal forecasting with a large ensemble of deep-learning weather prediction models},
  author={Weyn, Jonathan~A. and Durran, Dale~R. and Caruana, Rich and Cresswell-Clay, Nathaniel},
  journal={Journal of Advances in Modeling Earth Systems},
  volume={13},
  number={2},
  pages={e2021MS002502},
  year={2021}
}

@article{lang2024crps,
  title={AIFS-CRPS: ensemble forecasting using a model trained with a loss function based on the continuous ranked probability score},
  author={Lang, Simon and Alexe, Mihai and Clare, Mariana CA and Roberts, Christopher and Adewoyin, Rilwan and Bouall{\`e}gue, Zied Ben and Chantry, Matthew and Dramsch, Jesper and Dueben, Peter D and Hahner, Sara and others},
  journal={arXiv preprint arXiv:2412.15832},
  year={2024}
}

@article{chen2023fuxi,
  title={FuXi: a cascade machine learning forecasting system for 15-day global weather forecast},
  author={Chen, Lei and Zhong, Xiaohui and Zhang, Feng and Cheng, Yuan and Xu, Yinghui and Qi, Yuan and Li, Hao},
  journal={npj climate and atmospheric science},
  volume={6},
  number={1},
  pages={190},
  year={2023},
  publisher={Nature Publishing Group UK London}
}

@misc{buizza2015,
  title = {The Forecast Skill Horizon},
  author = {Buizza, Roberto and Leutbecher, Martin},
  year = {06/2015 2015},
  number = {754},
  doi = {10.21957/6g2wkoyb6},
  url = {https://www.ecmwf.int/node/8450},
}

@article{chen2025fengwu,
  title={The operational medium-range deterministic weather forecasting can be extended beyond a 10-day lead time},
  author={Chen, Kang and Han, Tao and Ling, Fenghua and Gong, Junchao and Bai, Lei and Wang, Xinyu and Luo, Jing-Jia and Fei, Ben and Zhang, Wenlong and Chen, Xi and others},
  journal={Communications Earth \& Environment},
  volume={6},
  number={1},
  pages={518},
  year={2025},
  publisher={Nature Publishing Group UK London}
}

@article{keisler2022gnn,
  title        = {Forecasting Global Weather with Graph Neural Networks},
  author       = {Keisler, Ryan},
  journal      = {arXiv preprint arXiv:2202.07575},
  year         = {2022}
}

@article{sun2025fuxiweather,
  title        = {A data-to-forecast machine learning system for global weather},
  author       = {Sun, Xiuyu and Zhong, Xiaohui and Xu, Xiaoze and Huang, Yuanqing and Li, Hao and Neelin, J. David and Chen, Deliang and Feng, Jie and Han, Wei and Wu, Libo and Qi, Yuan},
  journal      = {Nature Communications},
  volume       = {16},
  number       = {6658},
  pages        = {1--12},
  year         = {2025}
}

@article{zhong2024fuxiens,
  title        = {FuXi-Ens: A machine learning model for medium-range ensemble weather forecasting},
  author       = {Zhong, Xiaohui and Sun, Xiuyu and Xu, Xiaoze and Qi, Yuan and Neelin, J. David and Chen, Deliang and Li, Hao and Wu, Libo},
  journal      = {arXiv preprint arXiv:2405.05925},
  year         = {2024}
}

@article{lang2024aifs,
  title        = {AIFS: ECMWF's data-driven forecasting system},
  author       = {Lang, Simon and Ravuri, Shreya and Botev, Aleksandar and de B{\'e}zenac, Emmanuel and Ayed, Ismail and Lenc, Karel and Clark, Robin and Khan, Mufeed and Yang, Chan and Franklin, Robin and others},
  journal      = {arXiv preprint arXiv:2406.01465},
  year         = {2024}
}

@article{li2024generative,
  title        = {Generative emulation of weather forecast ensembles with diffusion models},
  author       = {Li, Liping and Carver, Ronald W. and Lopez-Gomez, Ignacio and Sha, Fei and Anderson, Joshua L.},
  journal      = {Science Advances},
  volume       = {10},
  number       = {6},
  pages        = {eadf8537},
  year         = {2024}
}

@article{documentation2020part,
  title={PART V: ENSEMBLE PREDICTION SYSTEM},
  author={DOCUMENTATION--Cy40r1, IFS},
  year={2020}
}

@article{hamill2022reanalysis,
  title={The reanalysis for the global ensemble forecast system, version 12},
  author={Hamill, Thomas M and Whitaker, Jeffrey S and Shlyaeva, Anna and Bates, Gary and Fredrick, Sherrie and Pegion, Philip and Sinsky, Eric and Zhu, Yuejian and Tallapragada, Vijay and Guan, Hong and others},
  journal={Monthly Weather Review},
  volume={150},
  number={1},
  pages={59--79},
  year={2022}
}

@article{li2021spatial,
  title={Spatial-temporal changes and associated determinants of global heating degree days},
  author={Li, Yuanzheng and Li, Jinyuan and Xu, Ao and Feng, Zhizhi and Hu, Chanjuan and Zhao, Guosong},
  journal={International journal of environmental research and public health},
  volume={18},
  number={12},
  pages={6186},
  year={2021},
  publisher={MDPI}
}

@article{wang2025short,
  title={Short-term residential electricity consumption forecast considering the cumulative effect of temperature, dual decomposition technology and integrated deep learning},
  author={Wang, Lanlan and Lin, Yong and Song, Tingting and Chen, Yuchun and Li, Kai and Ran, Junchao},
  journal={Energy Informatics},
  volume={8},
  number={1},
  pages={94},
  year={2025},
  publisher={Springer}
}

@inproceedings{liu2021swin,
  title={Swin transformer: Hierarchical vision transformer using shifted windows},
  author={Liu, Ze and Lin, Yutong and Cao, Yue and Hu, Han and Wei, Yixuan and Zhang, Zheng and Lin, Stephen and Guo, Baining},
  booktitle={Proceedings of the IEEE/CVF international conference on computer vision},
  pages={10012--10022},
  year={2021}
}

@inproceedings{hassani2023neighborhood,
  title={Neighborhood attention transformer},
  author={Hassani, Ali and Walton, Steven and Li, Jiachen and Li, Shen and Shi, Humphrey},
  booktitle={Proceedings of the IEEE/CVF conference on computer vision and pattern recognition},
  pages={6185--6194},
  year={2023}
}

@article{gorski2005healpix,
  title={HEALPix: A framework for high-resolution discretization and fast analysis of data distributed on the sphere},
  author={Gorski, Krzysztof M and Hivon, Eric and Banday, Anthony J and Wandelt, Benjamin D and Hansen, Frode K and Reinecke, Mstvos and Bartelmann, Matthia},
  journal={The Astrophysical Journal},
  volume={622},
  number={2},
  pages={759},
  year={2005},
  publisher={IOP Publishing}
}

@article{gneiting2007strictly,
  title   = {Strictly proper scoring rules, prediction, and estimation},
  author  = {Gneiting, Tilmann and Raftery, Adrian E},
  journal = {Journal of the American Statistical Association},
  volume  = {102},
  number  = {477},
  pages   = {359--378},
  year    = {2007}
}

@article{murphy1993good,
  title={What is a good forecast? An essay on the nature of goodness in weather forecasting},
  author={Murphy, Allan H},
  journal={Weather and forecasting},
  volume={8},
  number={2},
  pages={281--293},
  year={1993}
}

@book{katz1997economic,
  title={Economic value of weather and climate forecasts},
  author={Katz, Richard W and Murphy, Allan H},
  year={1997},
  publisher={Cambridge University Press}
}

@article{stock2025swift,
  title={Swift: An autoregressive consistency model for efficient weather forecasting},
  author={Stock, Jason and Arcomano, Troy and Kotamarthi, Rao},
  journal={arXiv preprint arXiv:2509.25631},
  year={2025}
}

@article{willard2025analyzing,
  title={Analyzing and exploring training recipes for large-scale transformer-based weather prediction},
  author={Willard, Jared D and Harrington, Peter and Subramanian, Shashank and Mahesh, Ankur and O’Brien, Travis A and Collins, William D},
  journal={Artificial Intelligence for the Earth Systems},
  volume={4},
  number={2},
  pages={240061},
  year={2025},
  publisher={American Meteorological Society}
}

@article{nguyen2025omnicast,
  title={OmniCast: A Masked Latent Diffusion Model for Weather Forecasting Across Time Scales},
  author={Nguyen, Tung and Pham, Tuan and Arcomano, Troy and Kotamarthi, Veerabhadra and Foster, Ian and Madireddy, Sandeep and Grover, Aditya},
  journal={arXiv preprint arXiv:2510.18707},
  year={2025}
}

@article{alexe2024graphdop,
  title={GraphDOP: Towards skilful data-driven medium-range weather forecasts learnt and initialised directly from observations},
  author={Alexe, Mihai and Boucher, Eulalie and Lean, Peter and Pinnington, Ewan and Laloyaux, Patrick and McNally, Anthony and Lang, Simon and Chantry, Matthew and Burrows, Chris and Chrust, Marcin and others},
  journal={arXiv preprint arXiv:2412.15687},
  year={2024}
}

@article{richardson2000skill,
  title={Skill and relative economic value of the ECMWF ensemble prediction system},
  author={Richardson, David S},
  journal={Quarterly Journal of the Royal Meteorological Society},
  volume={126},
  number={563},
  pages={649--667},
  year={2000},
  publisher={Wiley Online Library}
}

@article{raeth2025evaluating,
  title={Evaluating Weather Forecasts from a Decision Maker's Perspective},
  author={Raeth, Kornelius and Ludwig, Nicole},
  journal={arXiv preprint arXiv:2512.14779},
  year={2025}
}

\addtocontents{toc}{\protect\setcounter{tocdepth}{0}}
\appendix

\newpage

\section{Details on implementations and operational}

In all experiments, the assimilated history $X_{\le 0}$ and conditioning fields $C$ in the forecasting setup of Section~\ref{sec:decision} are taken from ERA5 reanalysis on the native $0.25^\circ$ grid, and diagnostic targets $Y_t$ (daily extrema, accumulations, rolling averages) are computed from the same ERA5 fields using the aggregation operators that define our emission heads (Sections~\ref{sec:typology} and~\ref{sec:gem}).

\subsection{Backbone of \methodtwo}

We have found that we can obtain improved model accuracy with little change in computational cost with \methodtwo. We do this by implementing an adapted variant of \emph{neighborhood attention} (NAT) \citep{hassani2023neighborhood}. NAT formulates the local inductive bias of weather dynamics continuously, unlike the discretized windows of Swin. As before, given gridded inputs
$x \in \mathbb{R}^{B \times C_{\text{in}} \times H \times W}$ and a stochastic noise vector $z \sim \mathcal{N}(0, I)$, the network realizes a deterministic mapping
\((\hat{X}_t, \hat{Y}_t) \;=\; f_\theta(X_{t-1}, C, z),
\) and different draws of $z$ induce an ensemble of samples $(\hat{X}_t^{(n)}, \hat{Y}_t^{(n)})$ from the same parameterization, as in \method. 

We adopt a two-stage attention design that separates pixel-level detail from coarse latent processing. Concretely, we run parallel attention at two resolutions:
\begin{itemize}[leftmargin=*]
    \item \emph{Pixel-resolution encoder/decoder blocks}, which operate directly on the $(H, W)$ grid. We have found that adding additional information (e.g. positional embeddings) within the pixel-space resolution helps performance. 
    \item A \emph{coarse latent processor}, which operates on a downsampled grid to capture large-scale organizing modes at reduced cost.
\end{itemize}

We designed \methodtwo to be a weather-adapted Neibghorhood attention backbone. There are three key design choices within each block:

\begin{algorithm}[h]
\caption{Boundary neighborhood attention}
\label{alg:boundary-nat}
\begin{algorithmic}[1]
\Require $x \in \mathbb{R}^{B \times C \times H_\text{grid} \times W_\text{grid}}$, neighborhood size $(k_h,k_w)$ with $k_w > k_h$
\Ensure Updated tensor $x' \in \mathbb{R}^{B \times C \times H_\text{grid} \times W_\text{grid}}$
\Statex \textit{Neighborhood construction}
\State For each grid point $(i,j)$, define an anisotropic $(k_h,k_w)$ neighborhood on the $(H,W)$ grid.
\Statex \textit{Boundary handling}
\State Apply periodic padding in longitude; wrap neighborhoods that cross left/right boundaries.
\State Apply non-periodic padding in latitude; mask entries that couple the two polar rows.
\Statex \textit{Projection and positional encoding}
\State Project $x$ to queries $Q$, keys $K$, values $V$ and apply 2D RoPE to $Q,K$ along latitude and longitude.
\Statex \textit{Local attention update}
\State For each $(i,j)$, compute attention restricted to its masked neighborhood and update features at $(i,j)$.
\State Collect all updated features into $x'$ and \Return $x'$.
\end{algorithmic}
\end{algorithm}

This backbone satisfies three properties within our modeling paradigm, i.e. (i) preserve geophysical inductive biases, (ii) remain computationally lightweight, and (iii) integrate with the probabilistic, single-shot training objective.

\paragraph{Computational Throughput and Latency.} 
GEFS produces 31-member 16-day forecasts every 6 hours using about 3 hours of wall-clock time under strict operational cutoffs. \textbf{FCN3} generates a 15-day global forecast in about one minute on a single GPU. \textbf{GraphCast} completes a 10-day forecast in less than one minute on one GPU. \textbf{FGN} requires approximately one minute per 15-day member on a TPU. \textbf{ArchesWeather} runs an efficient 1.5° coarse model with a 24-hour time step. \textbf{ArchesGen} produces a 15-day ensemble member in about one minute on an A100 GPU. \textbf{GenCast}, a diffusion model, takes about eight minutes per 15-day forecast at 0.25° resolution, which is too slow for four daily cycles or reforecasts.

\paragraph{Numerical and Autoregressive Stability.} 
\textbf{FCN3} demonstrates stable long-range rollouts up to 60 days with no drift. \textbf{FGN} maintains stable 15-day forecasts, with one outlier seed retrained through spectral quality control. \textbf{GraphCast} remains stable for 10-day predictions with no reported blow-up. \textbf{GenCast} provides stable 15-day ensembles with spectral energy distributions close to ERA5 truth, though diffusion iterations can drift if extended further. \textbf{ArchesWeather} achieves stable 15-day deterministic runs. \textbf{ArchesGen} delivers similarly stable probabilistic forecasts.

\subsection{On computational and inference time}
\label{sec:comp_inference_time}

We provide a quick comparison on training and inference times of popular weather models. 

Our primary model \method is trained in 19 H100 days for the swin architecture at 24-hour timestep resolution. This speedup of 22x to 100x at training time is driven by three primary structural choices: (a) the use of a Swin-style transformer which provides a 4-5x speedup over custom graph neural networks used in many other models. This speedup is partially explained by the efficient implementation of attention mechanisms; (b) the adoption of a native lat-lon grid that avoids the overhead of complex spherical meshes; and (c) a 24-hour temporal resolution that reduces the training sample volume by 4x. Furthermore, our sampling procedure is a one-shot predictor and therefore does not require denoising. The inference time gains relative to FGN are primarily due to the setup differences and not architectural differences: we use a 24-hour timestep instead of a 6-hour timestep (4x gains), and a single seed (4x further gains). 

\begin{figure}
    \centering
    \begin{subfigure}[t]{0.48\linewidth}
        \centering
        \includegraphics[trim={0.1cm 0.2cm 0.1cm 0},clip, width=\linewidth]{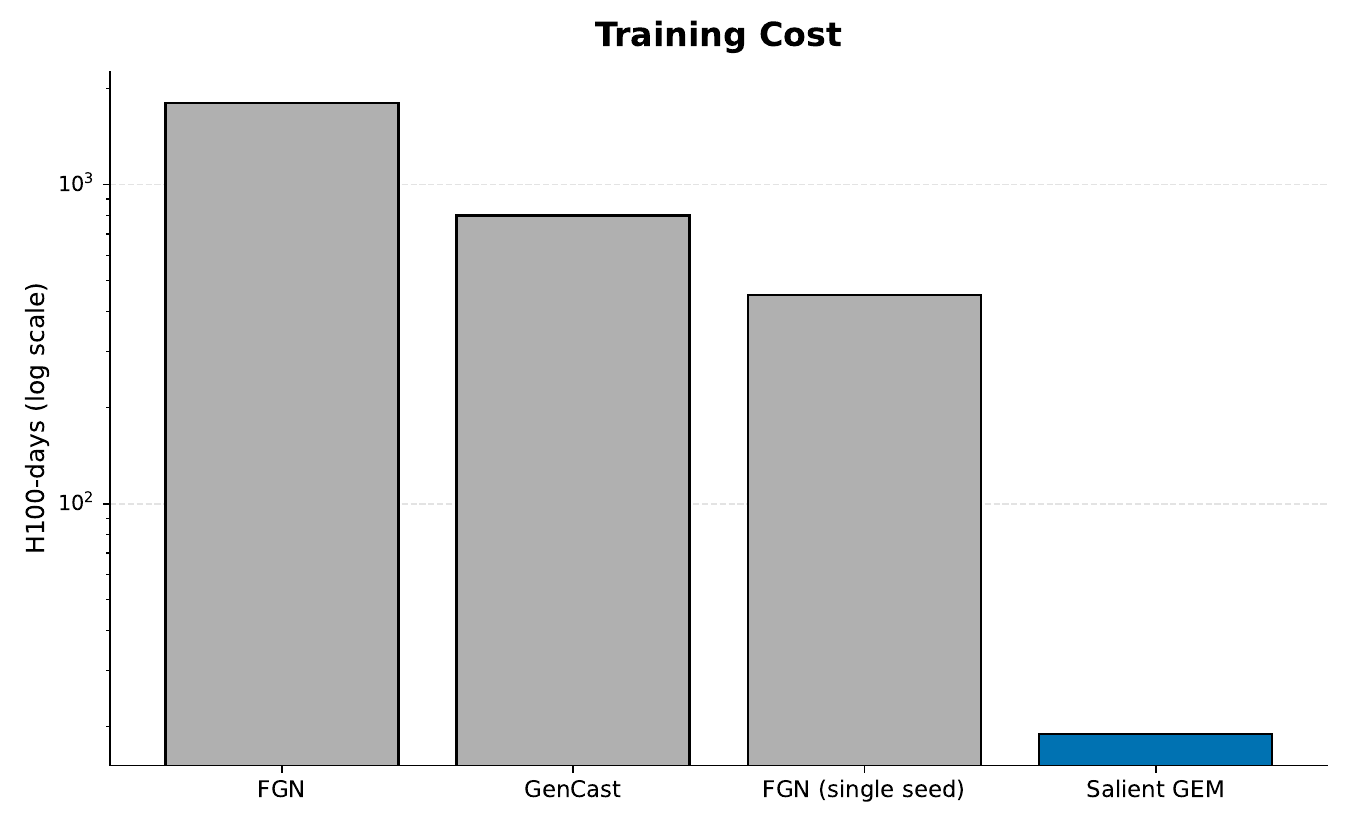}
        \caption{\textbf{Training costs.}}
        \label{fig:training_costs}
    \end{subfigure}
    \hfill
    \begin{subfigure}[t]{0.48\linewidth}
        \centering
        \includegraphics[trim={0.1cm 0.2cm 0.1cm 0},clip, width=\linewidth]{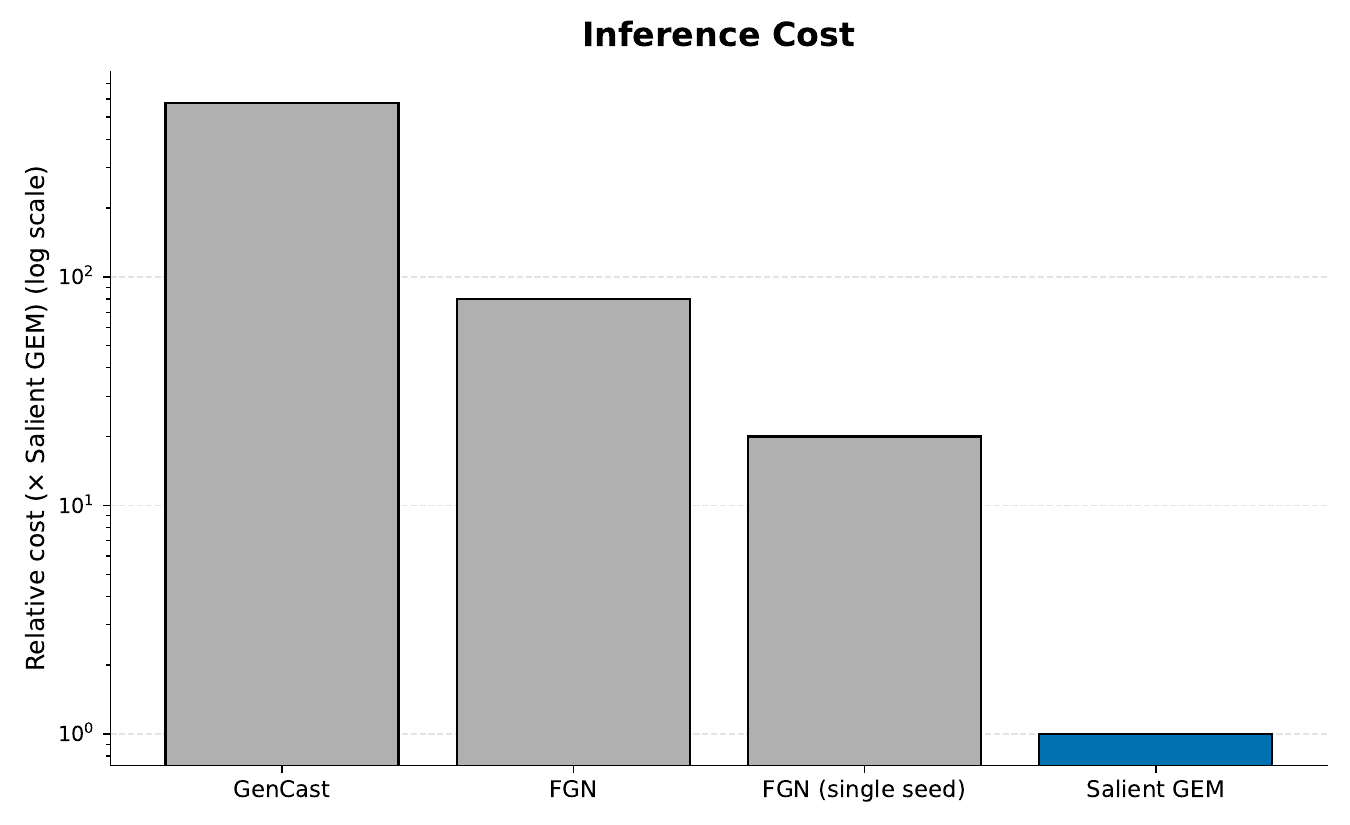}
        \caption{\textbf{Inference costs.}}
        \label{fig:inference_costs}
    \end{subfigure}
    \caption{\textbf{Computational costs across FGN and GenCast relative to GEM}. The design choices yield much faster training and inference costs relative to similar weather modeling approaches.}
    \label{fig:costs}
\end{figure}

\begin{table}[htbp]
\centering
\small
\begin{tabular}{p{2cm} p{4cm} p{3cm} p{3.5cm}}
\hline
\textbf{Model} & \textbf{Publicly stated training setup} & \textbf{H100-day estimate} & \textbf{Certainty of estimate} \\
\hline
FGN (WeatherNext 2) & 490 TPU v5p/v6e-days per model $\times$ 4 models & $\approx$1,800 H100-days ($\approx$450 per seed) & Good (directly from chip-days + FLOP ratios) \\
\hline
NeuralGCM & 3 weeks on 256 TPUs (finest-res model) & $\approx$3,000 H100-days & Moderate (assumes TPUs $\approx$ v4-class) \\
\hline
GenCast & No training budget given; diffusion on $0.25^\circ$ grid & $O(10^{2}\text{--}10^{3})$ H100-days, maybe $\sim10^{3}$ & Speculative (extrapolated from FGN / GraphCast scale) \\
\hline
\end{tabular}
\caption{Quick comparison of training compute estimates for weather forecasting models.}
\label{tab:compute_comparison}
\end{table}

\section{Additional results}

\subsection{Design choices \& Ablations}
Finally, we have discovered some minor artifacts in the predictions that were related to design choices of \method which we have addressed with \methodtwo. To perform and evaluate the sensitivity of various design choices, we use \methodtwo variant at $1^\circ$ resolution as a controlled test bed.
All ablations are trained with the same diagnostic-centric objective (pixel-space CRPS with optional spectral term) and daily time step. 

\textbf{Results}. We find that the architecture and setup is surprisingly stable across many ablations. Many common design choices, such as multi-lead fine-tuning, a spherical or Healpix grid, 2nd order Markov dynamics, among others, do not significantly impact the performance of our model. We find that only two changes do matter operationally. First, increasing the number of prognostic vertical levels from 3 to 7 to 13 consistently improves CRPS, with negligible inference-time cost because only the outer layers are affected. Second, adding the power spectrum loss term results in forecasts with better power spectra properties at higher frequencies.

Additional preliminary experiments (not shown) indicate that the relative insensitivity of the model to different ablation axes can be partially attributed to the coarse 24\,h timestep. By not explicitly modeling the diurnal cycle, the prognostic evolution of the atmosphere is significantly simplified and a major source of autoregressive error accumulation is removed. With sub-daily timesteps, tools like multi-lead fine-tuning become more important to maximizing model performance.

\begin{table*}[t]
\centering
\small
\renewcommand{\arraystretch}{1.25}
\setlength{\tabcolsep}{5pt}
\begin{tabularx}{\textwidth}{
  >{\hsize=1.3\hsize}X
  |>{\centering\arraybackslash\hsize=0.4\hsize}X
  >{\centering\arraybackslash\hsize=0.4\hsize}X
  >{\centering\arraybackslash\hsize=0.5\hsize}X
  |>{\hsize=1.6\hsize}X}
\toprule
\textbf{Ablation} & \textbf{Inputs} & \textbf{Training} & \textbf{Model} & \textbf{Effect on performance} \\
\midrule
Conditioning on diagnostics & \cmark & \xmark & \xmark & \xmark\; No effect. \\
\addlinespace
2nd order Markov & \cmark & \xmark & \xmark & \xmark\; Model degradation. \\
\addlinespace
Healpix grid & \cmark & \xmark & \xmark & \xmark\; No effect. \\
\addlinespace
Vertical Levels (3–37) & \cmark & \xmark & \xmark & \cmark\; Gains up to 13 levels. 37 levels destabilize long-leads \\
\hline
\addlinespace
Increasing \# samples ($2 \rightarrow 4$) & \xmark & \cmark & \xmark & \xmark\; No effect. \\
\addlinespace
Multi-lead rollout fine-tuning & \xmark & \cmark & \xmark & \xmark\; No effect. \\
Spherical loss & \xmark & \cmark & \xmark & \cmark\; Improves power spectrum \\
\hline
\addlinespace
Depth (8, 12, 18 layers) & \xmark & \xmark & \cmark & \xmark\; No effect.  \\
\addlinespace
Embedding dimension (768–1536) & \xmark & \xmark & \cmark & \xmark\; No effect. \\
\addlinespace
\bottomrule
\end{tabularx}
\caption{\textbf{Structured ablation matrix.} Each column indicates which component was modified (\cmark) or held fixed (\xmark). Only the vertical-level sweep produced a clear improvement (\cmark); all other perturbations yielded negligible or negative change.}
\label{tab:ablations_conditional}
\end{table*}

\paragraph{Summary.}
Across global scores, tail-focused metrics, S2S aggregates, and event-scale case studies, \method behaves as a decision-aligned probabilistic forecaster:
it delivers calibrated skill gains over operational ensembles, maintains useful subseasonal information while reverting to climatology, allocates probability mass sharply for high-impact events, and remains robust to most architectural perturbations.

\begin{figure}[!htbp]
    \centering
    \includegraphics[width=\linewidth]{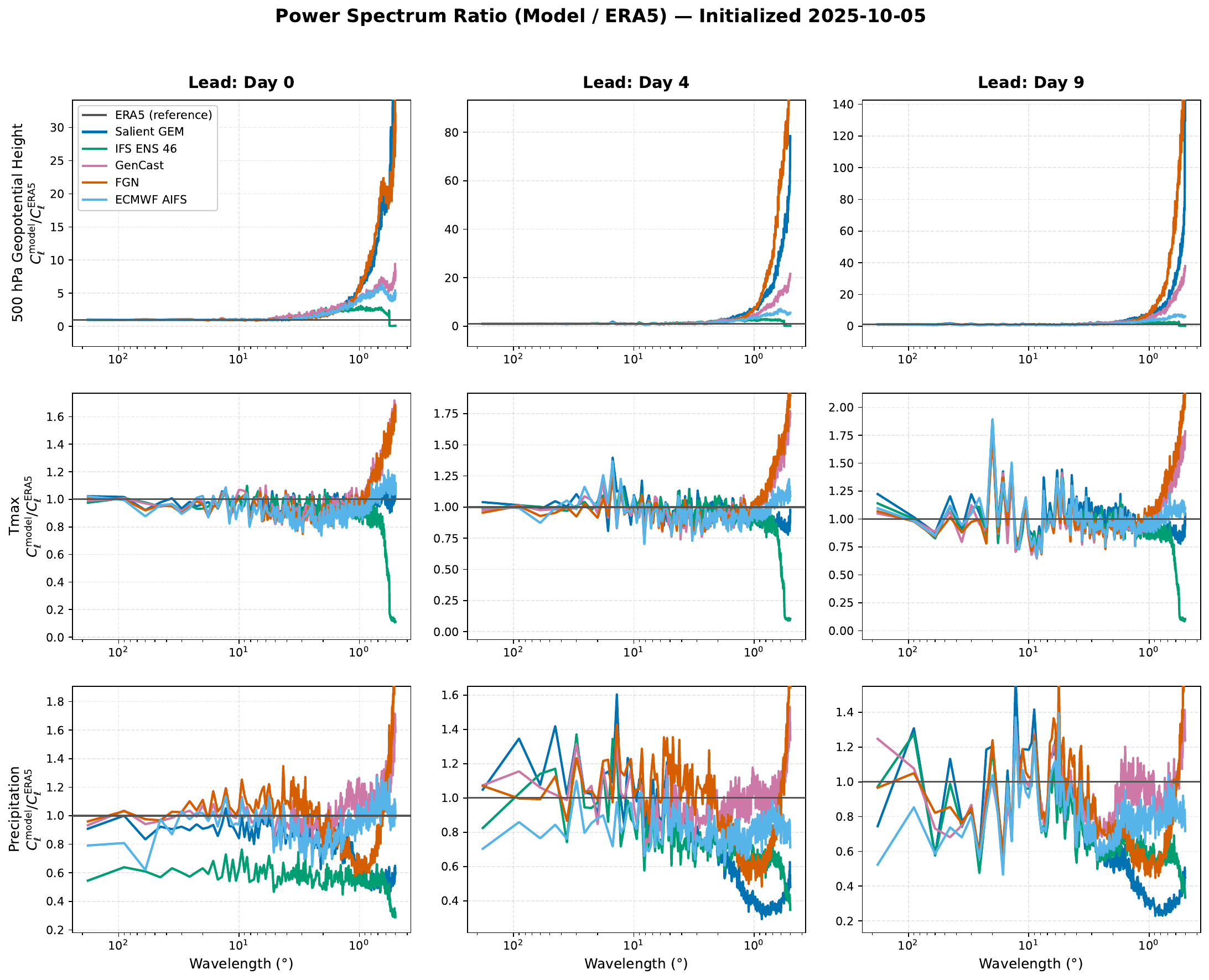}
\caption{\textbf{Power spectrum ratio (model / ERA5)} for 500 hPa geopotential height, maximum daily temperature, and precipitation at different lead times.
Each global field is expanded into spherical harmonics $Y_{\ell m}$ to compute the power spectrum $C_\ell = (2\ell+1)^{-1} \sum_m |a_{\ell m}|^2$, which quantifies variance as a function of spatial scale. Plotted is the ratio of model $C_\ell$ to ERA5 $C_\ell$, with wavelength $\lambda = 180°/\ell$ on the x-axis. A ratio of 1 indicates the model reproduces observed variance at that scale; >1 indicates excess power (too noisy), <1 indicates smoothing. Each curve is the mean ratio across 20 ensemble members for a single forecast initialized on 2025-10-05. The maximum degree $\ell_{\max}=359$ is set by the grid resolution.
}   \label{fig:power_spectra}
\end{figure}

\begin{figure}[!htbp]
    \centering
    \includegraphics[width=1\linewidth]{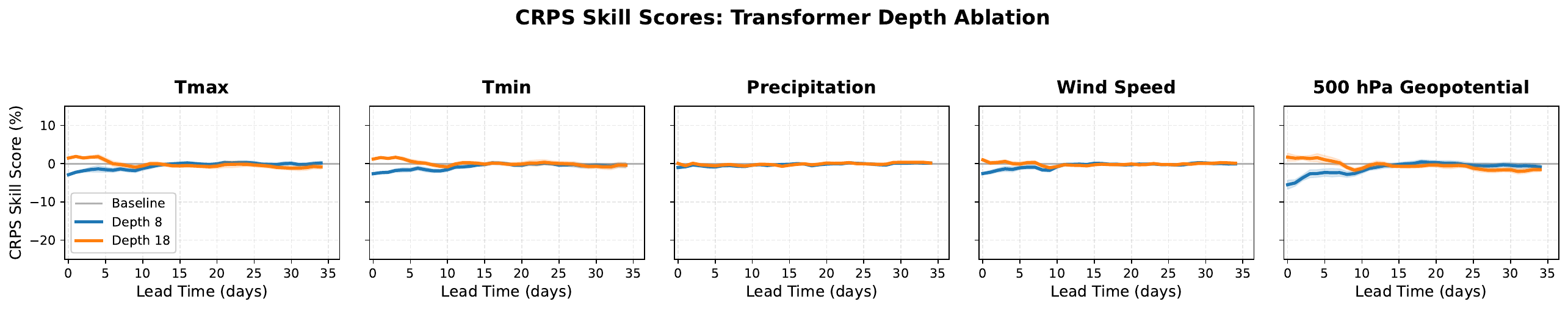}
    \caption{\textbf{Ablating the depth of the transformer}. For lead-times beyond 15-days, for all variables of interest, we observe no difference in the depth of the transformer architecture. We find minor performance gains in short lead times with higher depth.}
    \label{fig:ablation_depth_skill_scores}
\end{figure}

\begin{figure}[!htbp]
    \centering
    \includegraphics[width=1\linewidth]{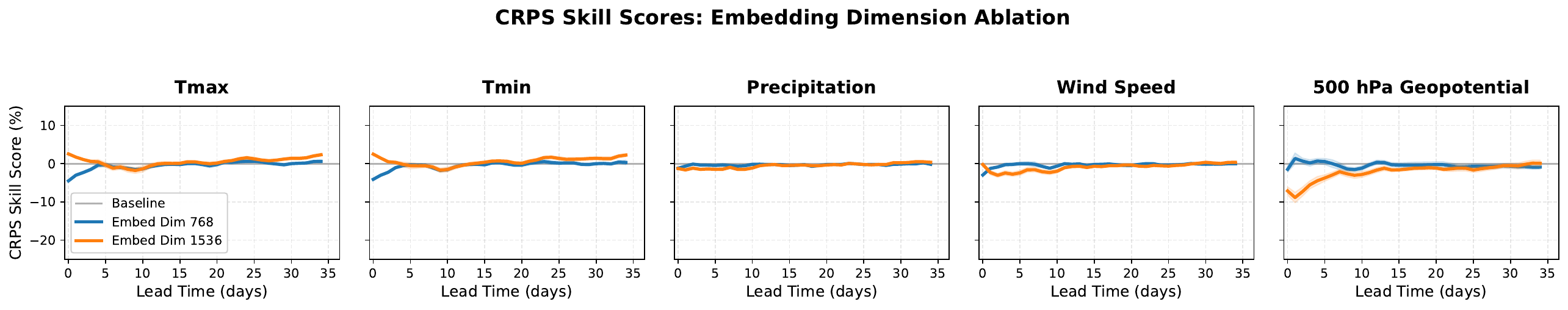}
    \caption{\textbf{Ablating the embedding dimension}. For lead-times beyond 10-days, we find no performance changes by changing the dimensionality of the embeddings. We find minor boosts in performance for the first couple of leads for some variables.}
    \label{fig:ablation_embed}
\end{figure}

\begin{figure}[!htbp]
    \centering
    \includegraphics[width=1\linewidth]{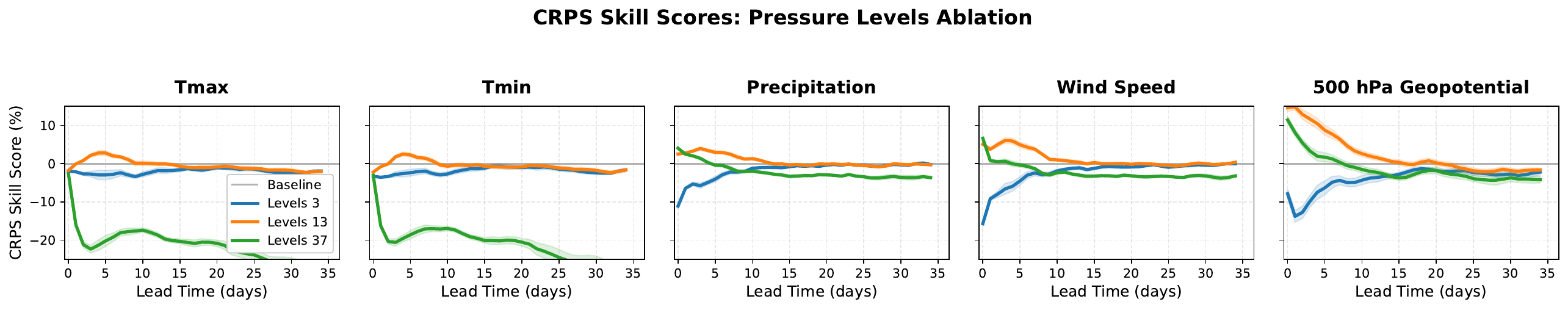}
    \caption{\textbf{Ablation by changing the number of pressure levels}. We find that increasing pressure levels from 3 to 13 increases the performance. Increasing pressure levels even further to 37 have an ambiguous effect and mostly deteriorate performance on variables of interest, e.g. Tmax and Tmin.}
    \label{fig:ablation_levels}
\end{figure}

\begin{figure}[!htbp]
    \centering
    \includegraphics[width=1\linewidth]{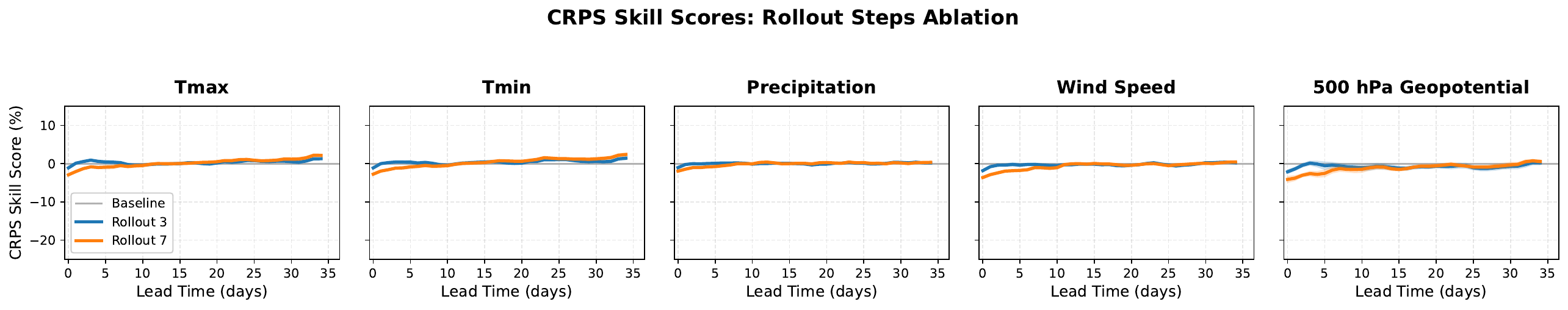}
    \caption{\textbf{Ablation by changing the number of rollout steps}. We do not find any consistent effect on increasing the rollout depth. In shorter lead-times, this tends to degrade performance. Longer lead-time behavior is variable dependent but largely has no effect on downstream performance. Baseline is a rollout length of one lead time. Backpropagation is performed with equal gradient contribution from each time. Additional mechanisms of balancing gradient contributions from different lead times, such as manual balancing or employing uncertainty-aware balancing, did not yield improvements over a single rollout length.}
    \label{fig:ablation_rollout}
\end{figure}

\begin{figure}[!htbp]
    \centering
    \includegraphics[width=0.9\textwidth]{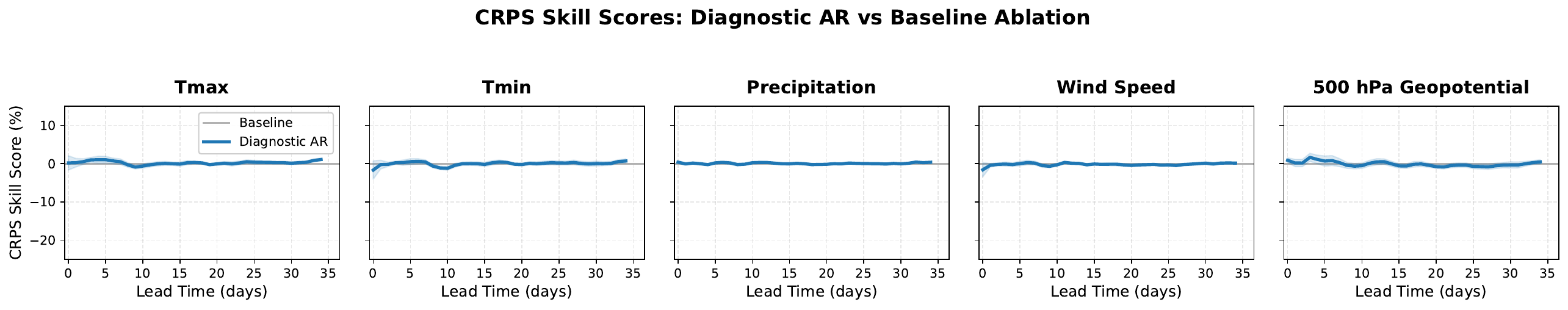}
    \caption{\textbf{Effects of including  diagnostic variables as a part of the conditioning put}. We find that, as expected, the performance of the model is not affected by conditioning on diagnostic outputs.}
    \label{fig:conditioning_vars}
\end{figure}

\begin{figure}[!htbp]
    \centering
    \includegraphics[width=0.9\textwidth]{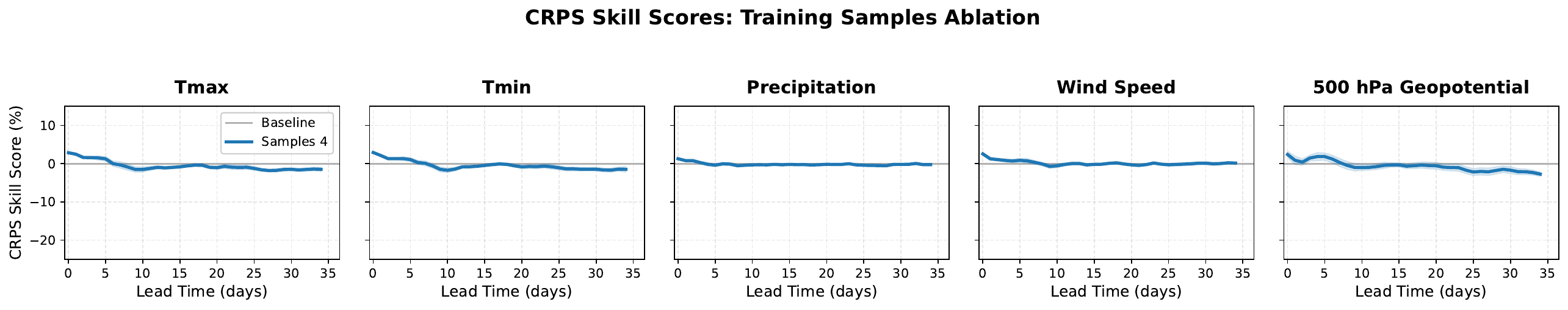}
    \caption{\textbf{Effects of increasing the number of CRPS samples}. We find that sampling two items at inference time is sufficient; sampling any further does not yield any performance gains but results in higher training costs.
}
    \label{fig:ablation_samples}
\end{figure}

\begin{figure}[!htbp]
    \centering
    \includegraphics[width=0.9\textwidth]{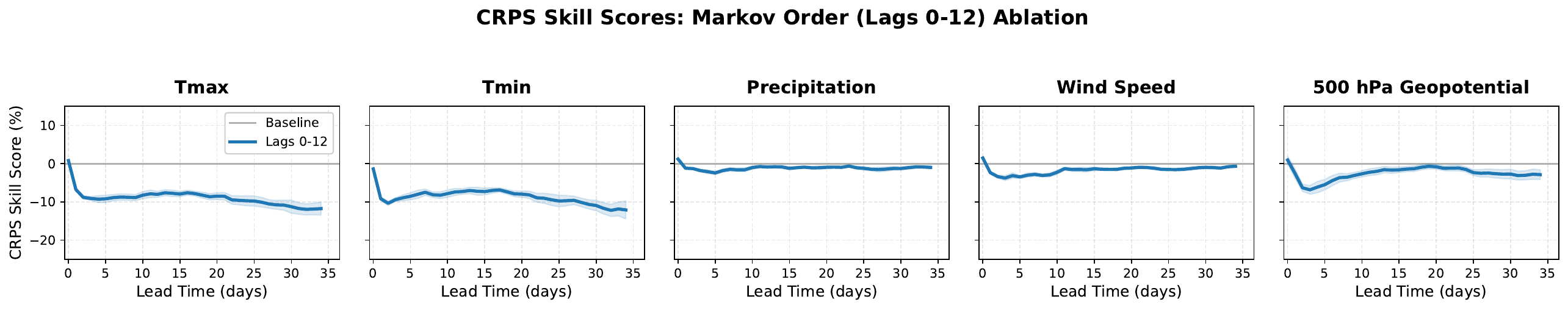}
    \caption{\textbf{Evaluating the effect of Markov order on downstream performance}. We find that relative to the baseline (1st order Markov assumption), moving to a 2nd order 2-in 2-out configuration, where we use two 12-hourly snapshots to evolve the prognostic fields, resulted in a net decrease in performance.}
    \label{fig:markov}
\end{figure}

\begin{figure}[!htbp]
    \centering
    \includegraphics[width=0.9\textwidth]{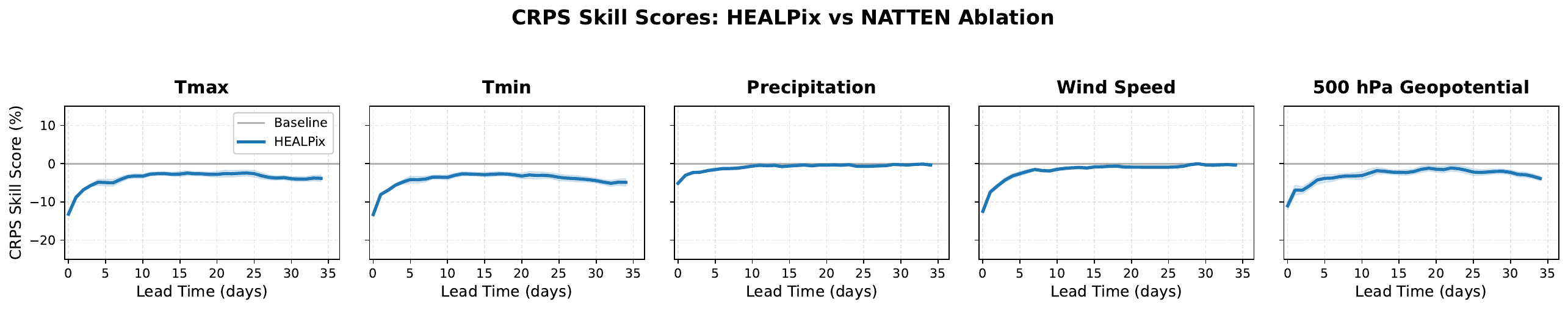}
    \caption{\textbf{Evaluating the effects changing subspaces on downstream performance.} We move to the Healpix grid \citep{gorski2005healpix} as our ablation. We find that the performance is either consistent with our rectangular grid or results in a net decrease in performance. Scores were evaluated on the rectangular grid, which may disadvantage the Healpix model due to accumulation of regridding errors.}
    \label{fig:healpix}
\end{figure}

\subsection{\method vs GEM 2.1}

\begin{figure}[!htbp]
    \centering
    \includegraphics[width=1\linewidth]{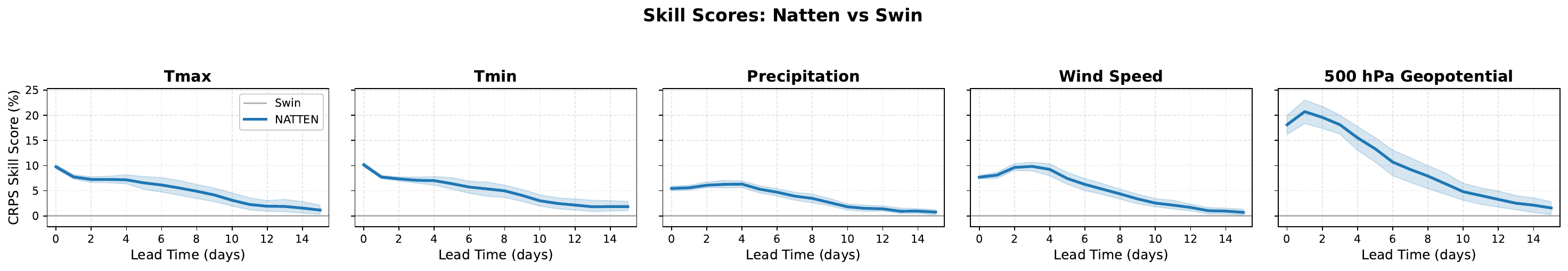}
    \caption{\textbf{Ablation on Natten vs Swin architecture}. We find that employing neigbhorhood attention (Natten), adopted in \methodtwo, leads to CRPS improvement over the Swin architecture which we employed in \method. }
    \label{fig:natten_vs_swin_skill_scores}
\end{figure}

\subsection{Spatial artifacts}

\begin{figure}[!htbp]
    \centering
    \includegraphics[width=1\linewidth]{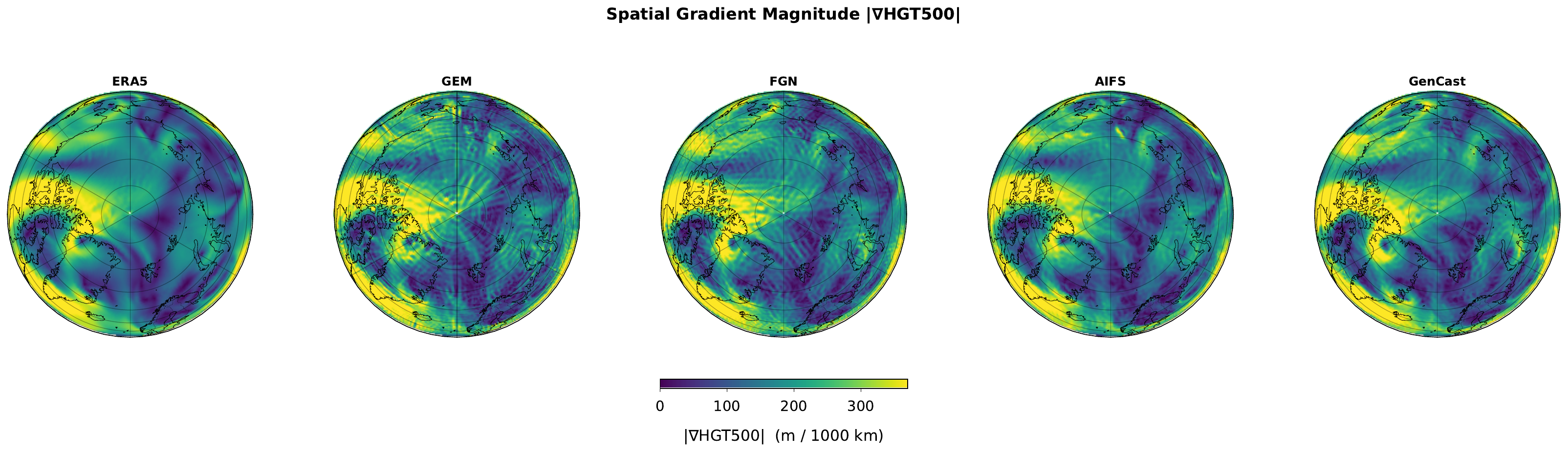}
\caption{\textbf{Illustration of grid-scale artifacts in different forecast models.} The gradient magnitude $|\nabla Z_{500}| = \sqrt{(\partial Z/\partial \theta)^2 + (1/\sin\theta \cdot \partial Z/\partial \phi)^2}$ is computed using spherical harmonic decomposition, which provides grid-independent derivatives and reveals artifacts arising from each model's native discretization. The north polar view emphasizes high-latitude regions where grid effects are most pronounced. GEM's regular lat-lon grid produces characteristic radial ``spokes'' near the pole due to meridian convergence. FGN and GenCast, which use icosahedral grids, exhibit faint hexagonal patterns. AIFS's reduced Gaussian grid shows the smoothest polar gradients. ERA5 reanalysis serves as reference for physically meaningful gradient structures.}

    \label{fig:gradient_hgt500}
\end{figure}

\clearpage

\subsection{Station validation}
\label{sec:station_validation}

\begin{figure}[!htbp]
    \centering
    \includegraphics[width=\linewidth]{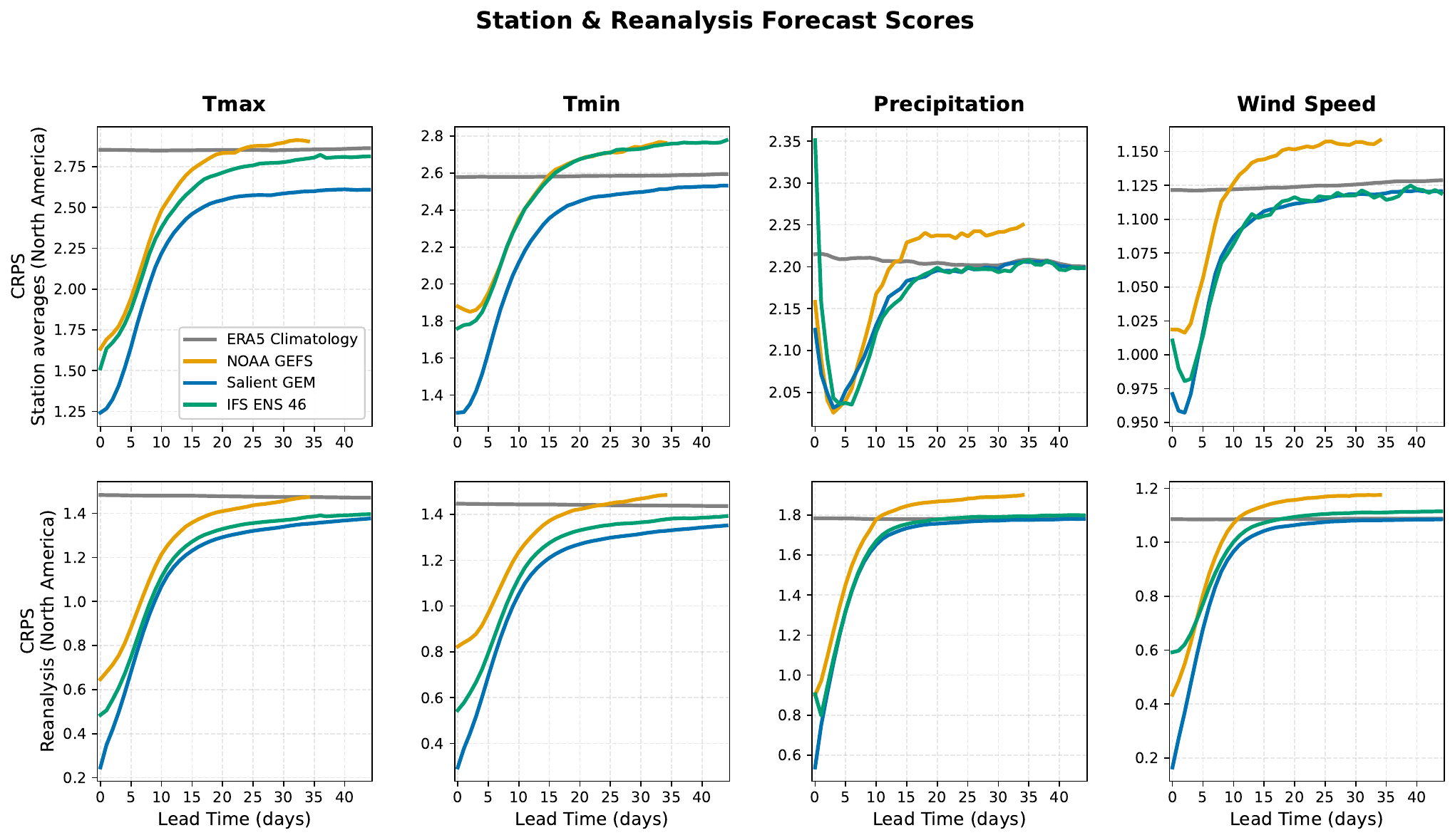}
\caption{\textbf{Absolute CRPS scores evaluated against GHCN station observations (top row) and reanalysis (bottom row)}. Station verification provides a ground-truth assessment independent of reanalysis, which models may be trained on. Station locations are shown in Figure~\ref{fig:station_locations}. \method achieves the lowest CRPS against station observations across all variables and lead times, with particularly strong performance for temperature forecasts. Scores were averaged over a period from 2023-06-30 to 2025-08-31.}
    \label{fig:station_absolute_scores_crps}
\end{figure}

\begin{figure}[!htbp]
    \centering
    \includegraphics[width=1\linewidth]{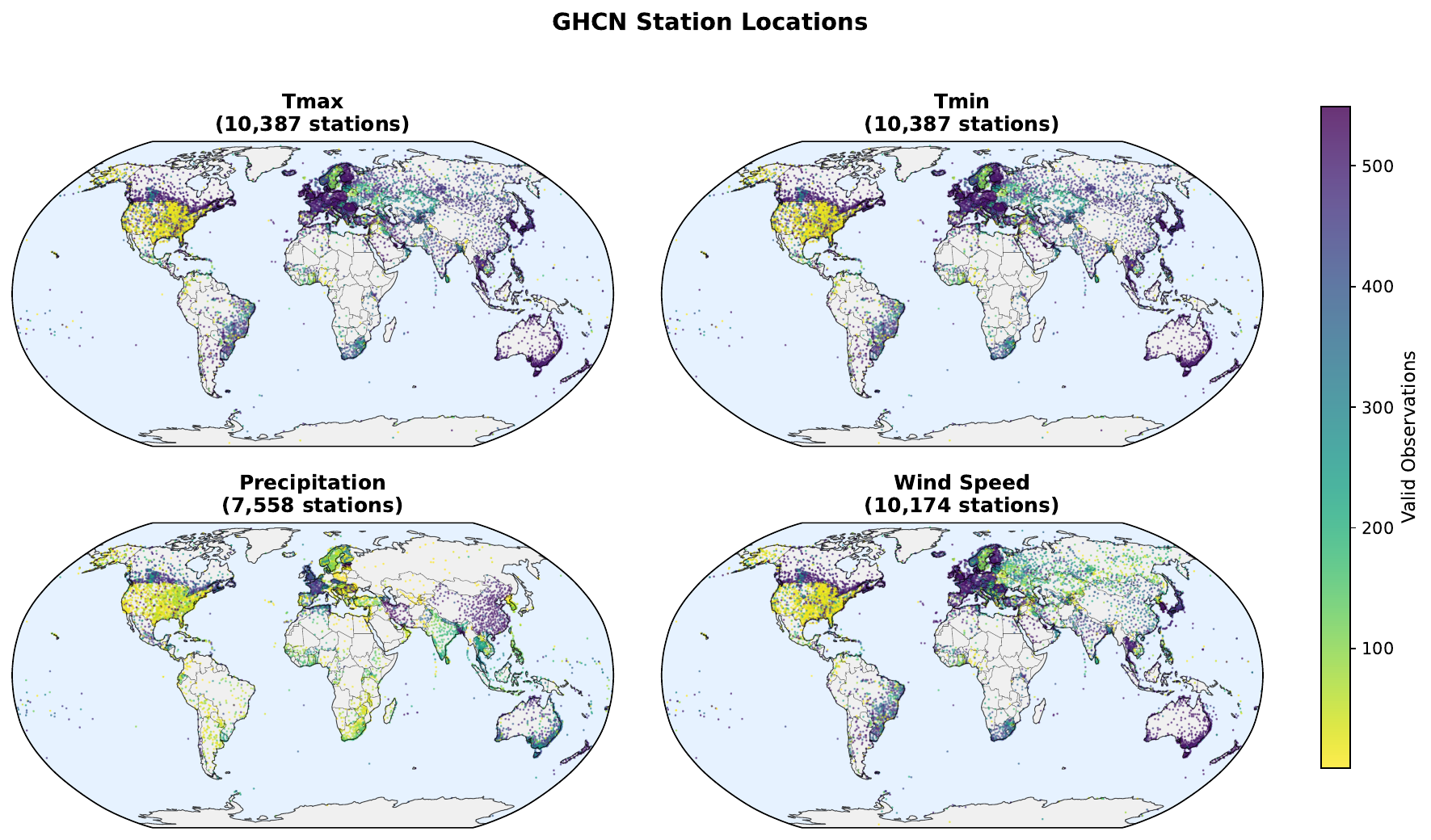}
    \caption{\textbf{GHCN-H station locations used for forecast verification.} Station observations are sourced from NOAA's Global Historical Climatology Network–Hourly (GHCN-H), which provides hourly meteorological measurements from the Integrated Surface Database (ISD). Raw hourly observations are aggregated to daily resolution, requiring $\geq$20 valid hours per day to ensure representative daily statistics. The data undergo a strict multi-stage quality control pipeline including: (1) ISD measurement flag filtering, (2) statistical outlier detection against co-located ERA5 reanalysis, (3) temporal spike detection, and (4) persistence checks. Each panel shows the geographic distribution of stations with valid observations for a given variable, with color indicating the number of valid daily observations per station during the evaluation period (2024).}
    \label{fig:station_locations}
\end{figure}

\clearpage

\subsection{Estimating daily extremes from sparse temporal snapshots}
\label{subsec:daily_extremes}
\begin{figure}[!htbp]
    \centering
    \includegraphics[width=0.8\linewidth]{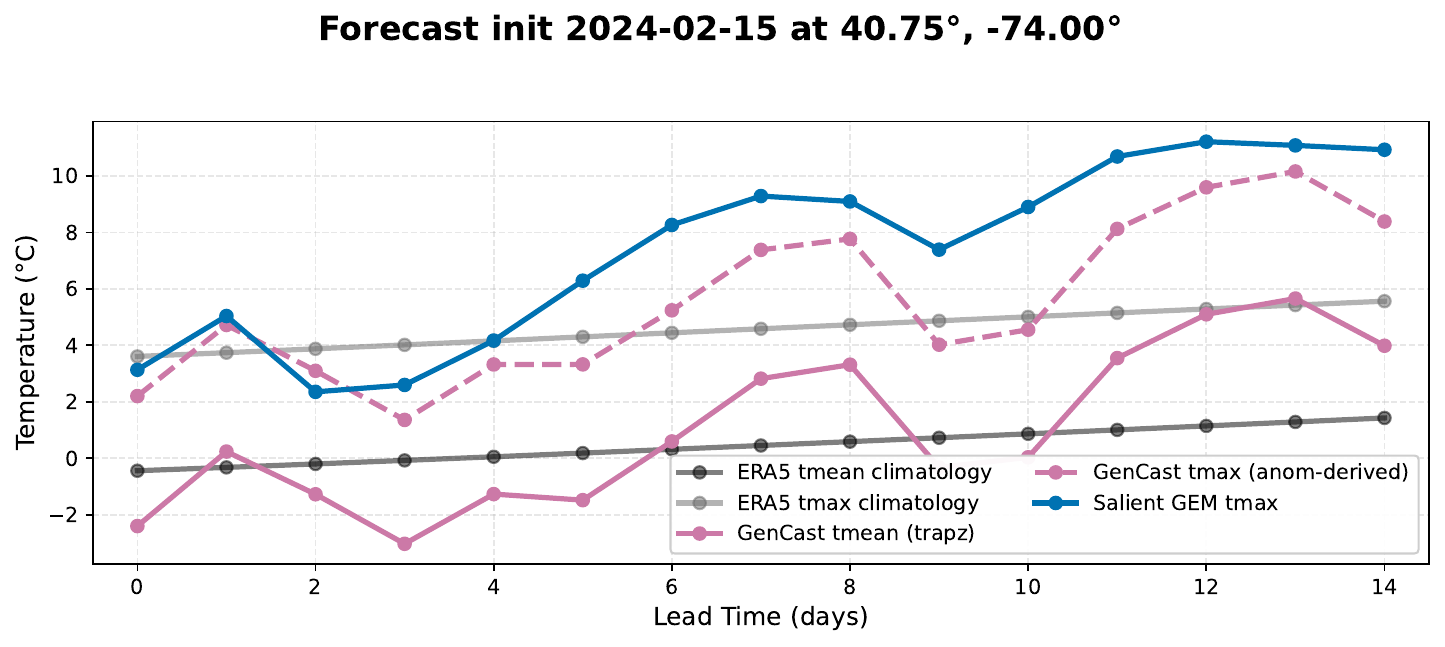}
    \caption{\textbf{Tmin/tmax derivation for ML models with sparse diurnal sampling}. GenCast outputs 12-hourly snapshots that miss daily extremes; we compute tmean via trapezoidal quadrature (solid pink), then add ERA5 climatological diurnal offsets to derive tmax (dashed pink). \method forecasts tmax natively (blue). This derivation is applied to all ML models with sparse sampling of the diurnal cycle.}
    \label{fig:tmin_tmax_deviation}
\end{figure}

Where models do not natively output extreme values (e.g. min/max outputs), we approximate them via a trapezoidal quadrature. An example is shown in Fig. \ref{fig:tmin_tmax_deviation} for a forecast initialization of 2024 February 15th.

\clearpage

\subsection{Scorecards for additional variables}
\begin{figure}[!htbp]
    \centering
    \includegraphics[width=1\linewidth]{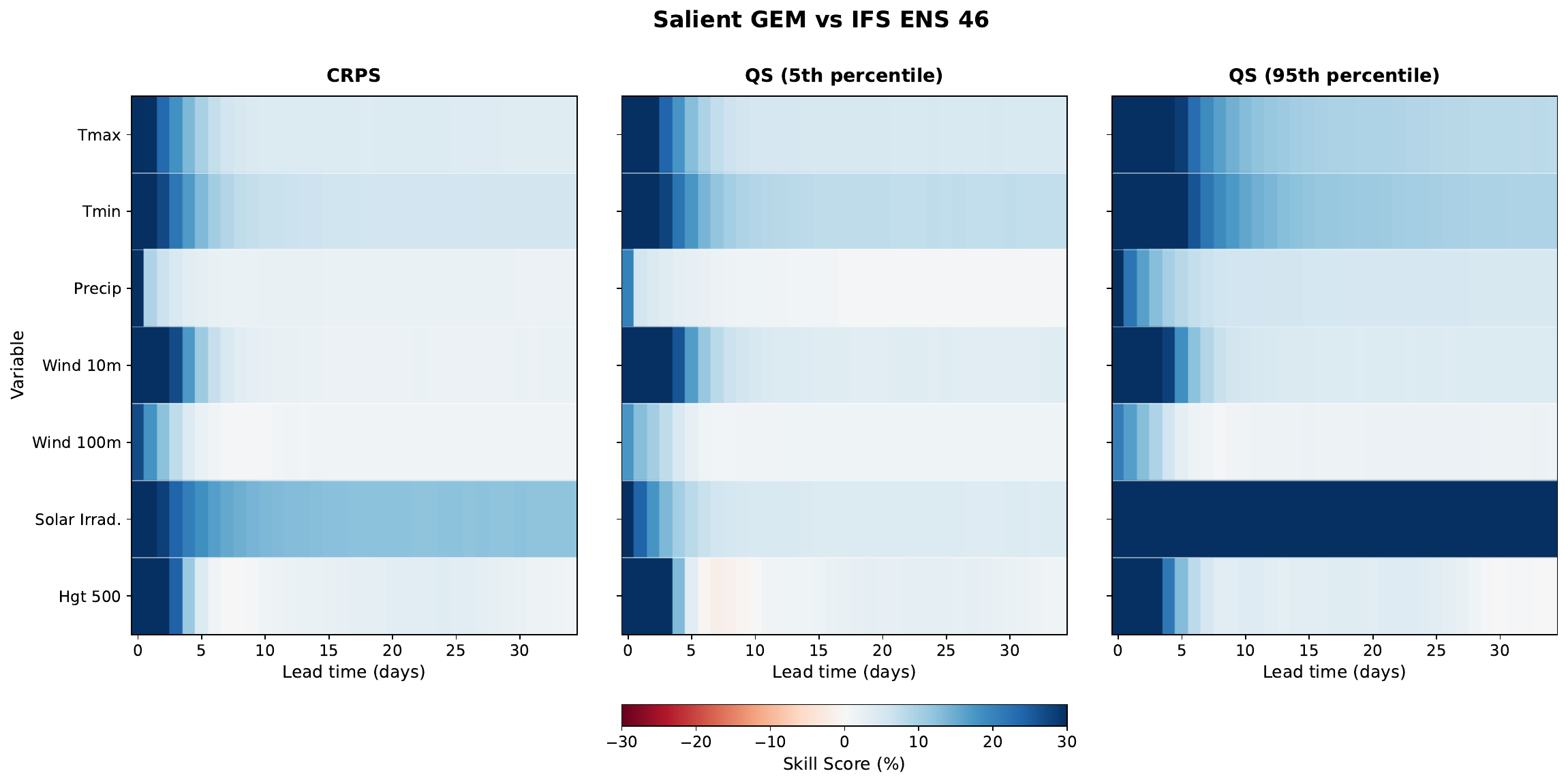}
    \caption{\textbf{\method vs ENS on chosen variables for CRPS and 5th and 9th percentiles}. We show improvements in skill score against IFS ENS across most lead times for the variables of interest.}
    \label{fig:multi_gem_vs_ens_domain}
\end{figure}

\begin{figure}[!htbp]
    \centering
    \includegraphics[width=1\linewidth]{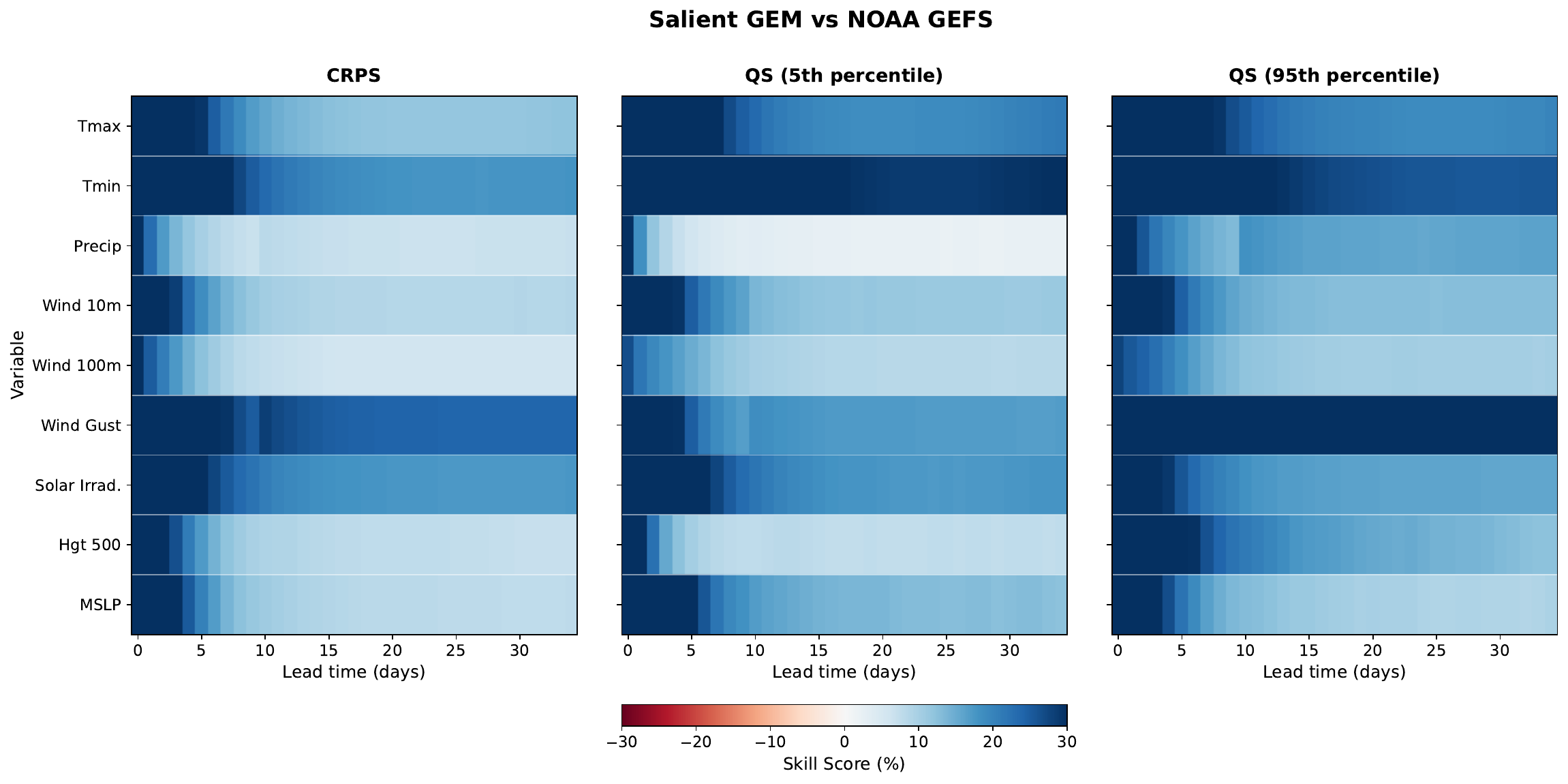}
    \caption{\textbf{\method vs NOAA GEFS on chosen variables for CRPS and 5th and 9th percentiles}. We show improvements in skill score against GEFS across most lead times for the variables of interest.}
    \label{fig:scorecard_multi_gem_vs_gefs_domain}
\end{figure}

\subsection{Relative Economic Value}
\label{subsec:rev_appendix}

\begin{figure}[!htbp]
    \centering
    \includegraphics[width=1\linewidth]{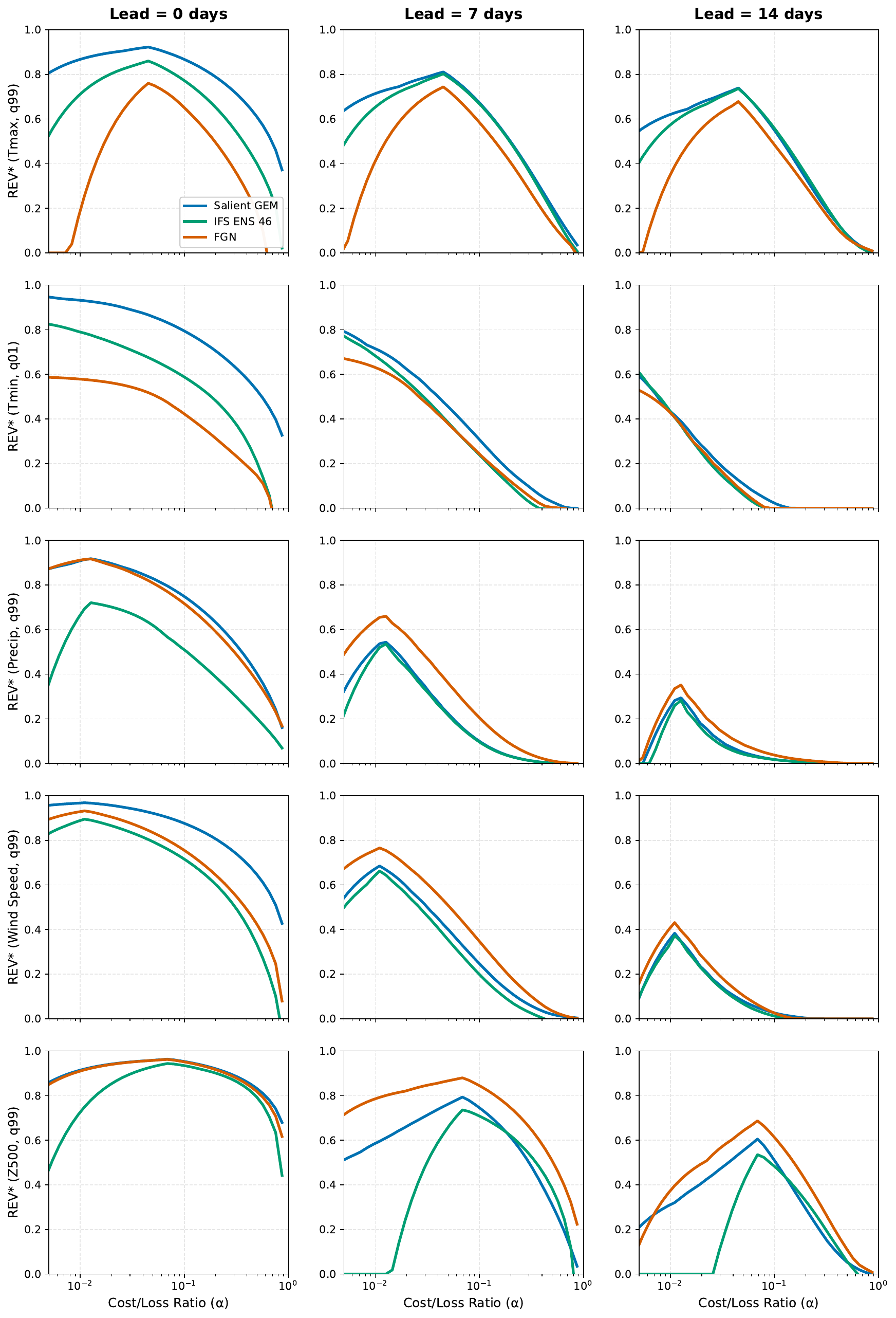}
    \caption{\textbf{Relative Economic Value (REV) calculations}. This grid of REV curves illustrates the comparative economic utility of three weather forecasting models—Salient GEM (blue), IFS ENS 46 (green), and FGN (orange)—across five meteorological variables and three lead times (0, 7, and 14 days). A universal trend is visible where forecast value degrades as lead time increases: the curves flatten and shift downward from left to right, indicating that all models provide less economic value at longer horizons due to increased uncertainty.}
    \label{fig:skill_scores_REV}
\end{figure}

\end{document}